%% file: main.tex
\documentclass{article}

    \PassOptionsToPackage{square,sort,comma,numbers}{natbib}

    \usepackage[preprint]{neurips_2021}

\usepackage[utf8]{inputenc} 
\usepackage[T1]{fontenc}    
\usepackage{url}            
\usepackage{booktabs}       
\usepackage{amsfonts}       
\usepackage{nicefrac}       
\usepackage{microtype}      
\usepackage[dvipsnames]{xcolor}         

\usepackage{amsmath}
\usepackage{amssymb}
\usepackage{relsize}
\usepackage{exscale}
\usepackage{mathabx}
\usepackage{booktabs}
\usepackage{graphicx}
\usepackage{xspace}
\usepackage{multirow}
\usepackage{pifont}

\newcommand{\modelname}{\textsc{VIMPAC}\xspace}

\newcommand{\ourmask}{mask-then-predict\xspace}
\newcommand{\ourcl}{contrastive learning\xspace}
\newcommand{\htm}{HowTo100M\xspace}

\newcommand{\ucf}{UCF101\xspace}
\newcommand{\hmdb}{HMDB51\xspace}
\newcommand{\ssv}{SSV2\xspace}
\newcommand{\ksmall}{Kinetics-400\xspace}

\newcommand{\diving}{Diving48\xspace}

\newcommand\fref{Fig.~\ref}
\newcommand\tref{Table~\ref}
\newcommand\sref{Sec.~\ref}

\definecolor{citecolor}{RGB}{34,139,34}
\usepackage[pagebackref=true,breaklinks=true,colorlinks,
citecolor=citecolor,bookmarks=false]{hyperref} 

\def\x{$\times$} 

\usepackage{wrapfig}

\usepackage{pifont}
\newcommand{\cmark}{\ding{51}}%
\newcommand{\xmark}{\ding{55}}%

\title{\modelname{}:  Video Pre-Training via Masked Token Prediction and Contrastive Learning}

\author{%
  Hao Tan\thanks{Equal contribution.}\,\,\,$^1$ \quad Jie Lei\footnotemark[1]\,\,\,$^1$ \quad Thomas Wolf\,$^2$ \quad Mohit Bansal\,$^1$ \\
  $^1$UNC Chapel Hill \quad $^2$Huggingface\\
  \texttt{\{haotan, jielei, mbansal\}@cs.unc.edu; thomas@huggingface.co} \\
}

\begin{document}

\maketitle

\begin{abstract}
Video understanding relies on perceiving the global content and modeling its internal connections (e.g., causality, movement, and spatio-temporal correspondence).
To learn these interactions, we apply a \ourmask  pre-training task on discretized video tokens generated via VQ-VAE.
Unlike language, where the text tokens are more independent, neighboring video tokens typically have strong correlations (e.g., consecutive video frames usually look very similar), and hence uniformly masking individual tokens will make the task too trivial to learn useful representations. 
To deal with this issue, we propose a block-wise masking strategy where we mask neighboring video tokens in both spatial and temporal domains.
We also add an augmentation-free contrastive learning method to further capture the global content by predicting whether the video clips are sampled from the same video.
We pre-train our model on uncurated videos and show that our pre-trained model can reach state-of-the-art results on several video understanding datasets (e.g., \ssv, \diving).
Lastly, we provide detailed analyses on model scalability and pre-training method design.\footnote{Code is released at \href{https://github.com/airsplay/vimpac}{https://github.com/airsplay/vimpac}. The pretrained checkpoints and scripts will be soon open-sourced in HuggingFace transformers.}

\end{abstract}

\input{1_intro}
\input{2_related}

\input{3_method}

\input{4_model}

\input{5_setup}

\input{6_result}

\input{7_analysis}

\section{Conclusion}
We presented the video pre-training framework \modelname that introduces \ourmask task to video self-supervised learning.
\ourmask task helps model spatio-temporal interactions that is important for video understanding.
We used the VQ-VAE quantizer and propose the block masking method that is essential to overcome the strong locality in video.
The \ourcl task is also added to learn separable global features.
Different from previous methods, our \ourcl did not use data augmentation over raw frames and is less sensitive to the temporal sampling distribution for positive pairs.
We showed that our frameworks could achieve state-of-the-art performance on two temporally-heavy dataset (\ssv and \diving) and reach competitive results on other datasets.
Detailed analyses were provided regarding the model scalability and task design.

\section*{Acknowledgement}
This work was granted access to the HPC resources of IDRIS under the allocation 20XX-AD011011621R1 made by GENCI. This work is supported by DARPA MCS Grant N66001-19-2-4031, DARPA KAIROS Grant FA8750-19-2-1004, and ARO-YIP Award W911NF-18-1-0336. Additionally, Hao Tan is supported by Bloomberg Data Science Ph.D. Fellowship and Jie Lei is supported by Adobe Research Fellowship.

\bibliographystyle{plain}
\bibliography{reference}

\input{appendix}

\end{document}

%% file: 1_intro.tex
\section{Introduction}
In recent years, state-of-the-art self-supervised methods have been exploring different directions for pre-training images and text representations, with Contrastive Learning (CL) providing strong results for vision representation learning~\cite{oord2018representation, chen2020simple, he2020momentum, chen2020improved, tian2020makes}, and Language Modeling (LM) becoming the de-facto standard in natural language processing (NLP) pre-training~\cite{devlin2019bert,liu2019roberta,yang2019xlnet,lan2019albert}.
These two approaches are quite different from each other. A contrastive objective compares positive/negative examples at a coarse/sample level, focusing on global content (e.g., for image classification) while a token modeling objective predict missing tokens from context at a much finer/sub-sample level to model sequential and short range interactions between tokens (e.g. in text generation tasks).
Interestingly, video understanding 
naturally combines both types of requirements. 2D processing along the spatial dimensions of the video bears similarity to image processing, while 1D processing along the temporal dimension often involves modeling sequential events and short range coherence. 

Hence, in this work, we propose to combine both text and image representation learning approaches for improved video pre-training, taking advantage of recent advances in self-supervised methods of both fields.
We name our method as \modelname: VIdeo pre-training via Masked token Prediction And Contrastive learning. 
From language research, we adopt a `masked language model' pre-training objective~\cite{devlin2019bert}
where a model is trained to reconstruct local masked regions in images or videos. From the computer vision world, we borrow a contrastive learning objective, specifically the InfoNCE~\cite{oord2018representation} objective applied on positive/negative video samples. While the masked language model objective encourages models to learn low-level semantics and sequential interaction, the contrastive loss provide a supervision for the models to learn more global and separable 
representations that are useful for many downstream tasks (e.g., action classification~\cite{soomro2012ucf101,kuehne2011hmdb,carreira2017quo}). Combining both objectives allow to provide training signal covering complementary aspects of a video signal: while short range correlations can be predominantly modeled from the training signal of the mask-and-predict task, the contrastive learning objective can provide signal on a more coarse-grained global-context and semantic level.

However, unlike language and its compact vocabulary of discrete tokens,  videos are typically represented as RGB pixels in an almost continuous, high dimensional vector space.
Naively masking pixels in videos induces a prohibitive computation cost while also tending to over-emphasize local details.
To overcome these issues, we first tokenize input videos using the latent codes of a pretrained Vector Quantized-Variational Auto-Encoder (VQ-VAE)~\cite{van2017neural,ramesh2021zero} to encode them in smaller quantized representations on which a reconstruction model can then be trained with a masked token modeling objective.
In practice, we also discovered that models trained with a uniform random token masking strategy can fail to learn meaningful and useful visual representations as neighboring pixels may contain very similar and correlated content (in particular along the temporal frame axis),
making the task of predicting a randomly masked token from its visible neighbors easy. 
We therefore also introduce a block-masking scheme for videos by simultaneously masking video tokens in a contiguous 3D spatio-temporal block.
Reconstructing such an extended spatio-temporal cube requires performing long-range predictions, forcing the models to learn a more complex set of relations between the video tokens, resulting in better visual representations.

Our contrastive learning approaches also departs from previous work in several aspects. 
First, since we apply the contrastive objective on token-discretized video samples and in combination with the token modeling loss, we observe strong performance without requiring the usual extensive set of data augmentations~\cite{chen2020simple,chen2020improved, qian2020spatiotemporal, feichtenhofer2021large}.
Second, we are able to leverage positive clip pairs that are temporally distant from each other (can be as far as 400 seconds away), 
while previous work favors using positives within a shorter range (maximum 36 seconds for uncurated videos in~\cite{feichtenhofer2021large} or 10 seconds in~\cite{qian2020spatiotemporal}).

We evaluate the performances of our method \modelname on several video understanding datasets including two temporally-heavy tasks, \ssv and \diving on which it achieves state-of-the-art results with regard to both self-supervised and supervised pre-training works and a set of more spatial-heavy datasets (\ucf, \hmdb, and \ksmall) on which it achieve competitive results with regards to the literature.
Overall, taking advantage of VQ-VAE discretized video tokens, we present a method for self-supervised learning of video representations that combines two general streams of research in self-supervision: masked language modeling and contrastive learning.
Our contribution is 3-folds: 
($i$) We apply the mask-then-predict task to video understanding and introduce the use of block masking. 
($ii$) We propose a contrastive learning method which is able to achieve strong performance without spatial data augmentation.
($iii$) We empirically show that this method can achieve state-of-the-art results on several video classification datasets.
We also present comprehensive ablation studies to analyze the various aspects of our proposed approach.

%% file: 2_related.tex
\section{Related Work}

Unsupervised representation learning, with the promise of learning from large-scale unlabeled data, has drawn increasing attention in recent years, in both computer vision and natural language processing (NLP) communities. 
Most mainstream self-supervised methods can be categorized into three general directions: generative, denoising, and discriminative~\cite{chen2020simple,grill2020bootstrap,doersch2015unsupervised}. 

Generative and denoising methods seek to generate or reconstruct corrupted text/image/video tokens according to their empirical distributions. In generative and auto-regressive methods, next tokens are predicted given a causal context~\cite{chen2020generative, Oord2016ConditionalIG} while denoising methods seek to reconstruct corrupted or masked tokens given an extended context~\cite{devlin2019bert, raffel2019exploring}
For text, since the tokens (words or sub-words~\cite{sennrich2016neural, wu2016google}) are discrete and has relatively high entropy rate, language modeling has became the de-facto approach for pre-training models for most natural language tasks~\cite{ruderetal2019transfer}.
In the case of images, generative approaches often operate on pixel space~\cite{bertalmio2001navier,yu2018generative,kim2019deep, chen2020generative, Oord2016ConditionalIG}, which can be extremely expensive for larger input size like videos and has hence limited the widespread adoption of these methods.
Recently, discretizing images and videos with discrete variational auto-encoders (VQ-VAE), has been explored in compression and generative setups ~\cite{van2017neural, Razavi2019GeneratingDH, walker2021predicting, ramesh2021zero, yan2021videogpt}.
Such approaches avoid modeling pixel-level details and have enabled the use of generative models for images and videos~\cite{ walker2021predicting, ramesh2021zero}. Differing from these works, our framework investigates the use of such quantized representations in a denoising/reconstruction setup rather than generative, which has been shown in the NLP community to learn better representations~\cite{raffel2019exploring, devlin2019bert}.
Moreover, beyond simply applying MLM to the video tokens, we propose a block masking strategy to reduce the strong local correlation in neighboring video tokens.
This 3D block masking strategy is inspired from recent span-masking schemes~\cite{raffel2019exploring,joshi2020spanbert} for language modeling. 

The other direction of research, which our framework combines, is discriminative methods which start from the hypothesis that learning to reconstruct local details is not necessary for learning good visual representations. In some of these approaches, an objective is constructed around hand-crafted heuristics tasks like spatial arrangement, color, playback speed or frame order predictions~\cite{doersch2015unsupervised, zhang2016colorful, gidaris2018unsupervised, fernando2017self,lee2017unsupervised,wei2018learning, epstein2020oops, benaim2020speednet}.
Another line of discriminative approaches is \emph{contrastive learning} which aims at training a model to be able to recognize different views (e.g., different augmentation of images or different temporal samples of videos) of the same image or video, as a way to learn general representations~\cite{chen2020simple,he2020momentum,chen2020improved,grill2020bootstrap,caron2020unsupervised,feichtenhofer2021large}. 
This direction of research is reminiscent of sentence-order prediction tasks introduced in NLP~\cite{devlin2019bert, lan2019albert} with the goal of predicting whether two text sequences should be juxtaposed or not, an approach challenged in more recent literature~\cite{liu2019roberta, lan2019albert}.
In the present work, inspired by visual representation rather than text representation learning literature, we adapt the contrastive learning approach to video by training a model to differentiate pairs of clips from a single video from pairs of clips from disparate videos.

%% file: 3_method.tex
\begin{figure}[t]
\vskip 0.1in
\begin{center}
\includegraphics[
                width=0.99\columnwidth,
                ]{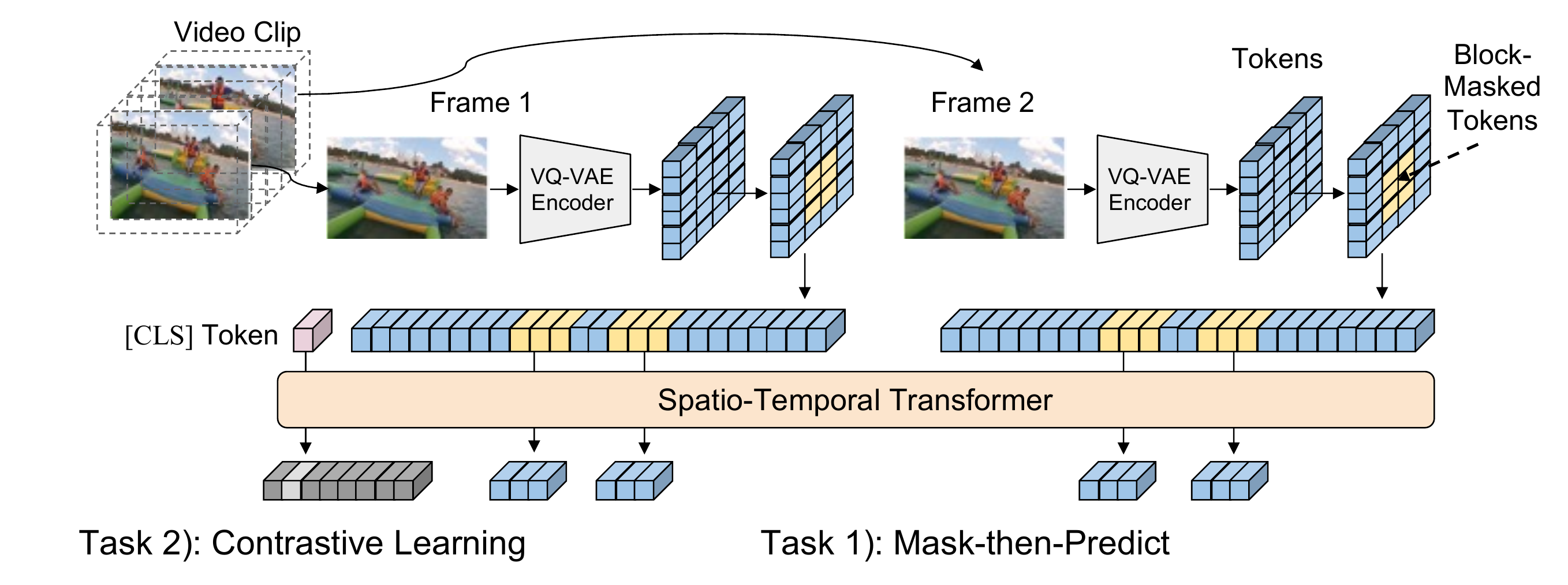}
                \vspace{-8pt}
\caption{
Illustration of our \modelname framework.
Frames are sampled from the video clip and discretized by VQ-VAE encoder. 
The tokens from VQ-VAE are then block-masked (in light yellow blocks).
The model is self-supervised by two tasks: 1) \emph{\ourmask} task predicts the masked tokens from visible context; 2) \emph{\ourcl} task classifies the positive examples (details in \fref{fig:block_mask_fig}) with the feature of the additional [CLS] token.
For space limit, we only show 2 frames and a smaller token map.
}
\label{fig:method_fig}
\end{center}
\vspace{-10pt}
\end{figure}

\section{Methods}
In this section, we present 
our proposed video pre-training method \modelname (illustrated in \fref{fig:method_fig}) as well its detailed components. 
We first introduce the \ourmask task in \sref{sec:mask_task}, and then the \ourcl task in \sref{sec:contrastive}. 
Lastly, we discuss how these two tasks are combined in \sref{sec:objective}.

\subsection{Mask-then-Predict Task}
\label{sec:mask_task}
Suppose that a video clip input comprises $T$ frames $\{f_1, f_2, \ldots, f_\textsc{t}\}$, 
the \ourmask task learns video representations by predicting the masked contents from their spatio-temporal context.
Denote the set of mask-token locations as $M$, we learn to predict the original tokens $\{x_{t, i, j}\}$ (see details below) by optimizing the negative log-likelihood:
\begin{align}
    \mathcal{L}_\text{mask} = - \frac{1}{\vert M \vert} \sum_{t, i, j \in \mathit{M}} \log p_{t, i, j} \left(x_{t, i, j} \mid \{x_{t', i', j'}\}_{t', i', j' \in M^\mathrm{C}}\right),
\end{align}
where $M^\mathrm{C}$ is the complement of $M$ and thus indicates the unmasked context.

\paragraph{Video Quantization with VQ-VAE.}
\label{sec:vqvae}
Since directly applying \ourmask over raw pixels and masking/predicting pixels leads to prohibitive computational costs and also tends to make the model overfit on detailed low-level visual information, 
we quantize the input videos with Vector Quantized-Variational Auto Encoder (VQ-VAE)~\cite{van2017neural,ramesh2021zero}.
The VQ-VAE encoder takes an image as input and produces a token map, where the tokens belong to a predefined vocabulary $V$ of cardinal `vocabulary size'. 
The VQ-VAE decoder then tries to reconstruct the original image from these latent codes.
In our method, we use a frozen and pretrained generic VQ-VAE encoder as a compressor that converts an input from an original input space $\mathbb{R}^{H \times W \times 3}$ into a discretized space $[V]^{\frac{H}{8} \times \frac{W}{8}}$.
We independently apply the VQ-VAE encoder to each frame $f_t$ inside a clip.\footnote{We do not use the Video-VQVAE~\cite{walker2021predicting} method since the image-trained VQVAE~\cite{ramesh2021zero} has been pretrained on a very large image corpus and as a consequence cover a much more diverse set of visual scenes and elements.}
We keep the VQ-VAE weights frozen and do not finetune or adapt this model on our corpus.

\begin{figure}[t]
\vskip 0.1in
\begin{center}
\includegraphics[
                width=0.99\columnwidth,
                ]{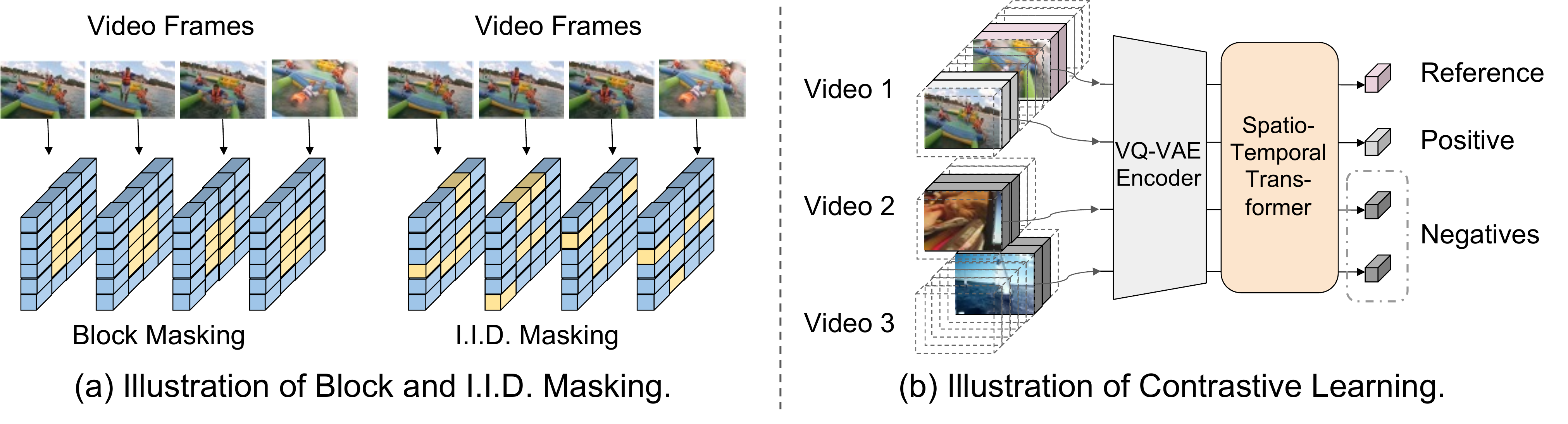}
                \vspace{-8pt}
\caption{
Illustration of pre-training task details. 
In (a), block masking constructs the 3D-contiguous masking cube instead of independently sampling masked tokens (i.e., i.i.d masking).
In (b), given the reference video clip, the positive clip is randomly sampled from the same video (video 1) while negative clips are sampled from other videos (video 2 and video 3). 
No spatial augmentations are applied to the raw video clips.
}
\label{fig:block_mask_fig}
\end{center}
\vspace{-10pt}
\end{figure}

\paragraph{Block Masking}
For sampling tokens to mask, the original BERT methods proposes the i.i.d. (independent and identically distributed) random mask $M_\text{iid}$ that constitutes of masked tokens:
\begin{align}
    \mathit{M}_\text{iid} =  \{(t, i, j) \mid \mathcal{U}_{t,i,j}[0,1] < \xi  \},
\end{align}
where $\mathcal{U}_{t,i,j}[0, 1]$ is the uniform distribution from $0$ to $1$ and $\xi$ is the threshold.
Intuitively, $\xi$ is the expectation of masked-token ratio and hence controls the difficulty of our \ourmask task.
In our early experiments, we found it easy to infer a masked token from its direct spatio-temporal neighbours (e.g., neighboring frames in a video tend to look similar thus contain similar tokens).
To overcome this issue, we propose to use block masking (see \fref{fig:block_mask_fig} (a)), which masks continuous tokens inside spatio-temporal blocks.
For each mask block $B$, we randomly sample lower ($B_{*, 0}$) and upper boundaries ($B_{*, 1}$) for each of the temporal ($T$), height ($H$), and width ($W$) 
dimensions.
The direct product of the intervals delimited by these boundaries constructs the block mask.
The final mask $\mathit{M}_\text{block}$ is the union of them:
\begin{align}
    \mathit{M}_\text{block} = \mathsmaller{\bigcup}_{B}  [B_{T, 0}, B_{T, 1}] \times [B_{H, 0}, B_{H, 1}] \times [B_{W, 0}, B_{W, 1}]. 
\end{align}

\subsection{Contrastive Learning}
\label{sec:contrastive}
Contrastive learning aims to distinguishing positive pairs from negative pairs (see \fref{fig:block_mask_fig} (b)).
For each video $\mathit{video}_i$, we uniformly and independently sample two clips $c_i$, $c_i'$ as a positive pair, while the clips in a batch belonging to other videos are used to construct negative pairs.
A model (described in \sref{sec:backbone}) processes clips $c_i$, $c_i'$ to build respective vector representations $f_i$, $f_i'$ and an InfoNCE~\cite{oord2018representation} loss is used to distinguishes the positive feature pair ($f_i$, $f_i'$) from the negative pairs $\bigcup \left\{\{(f_i, f_k), (f_i, f_k')\}  \mid k \neq i \right\}$ for each clip $c_i$:
\begin{align}
\mathcal{L}_\text{InfoNCE}(i) &= - \log \frac{\exp{(f_i^\top f_i' / \gamma)}}{\sum_{k\neq i} \exp{(f_i^\top f_k / \gamma)} + \sum_{k}\exp{(f_i^\top f_k' / \gamma)}},
\end{align}
which we combine with the symmetric loss $\mathcal{L}_\text{InfoNCE}'(i)$ for paired clip sample $c_i'$.

The final loss for a mini batch
$\mathcal{L}_\text{cl}$ is the average loss for all $n$ clips in the mini-batch:
\begin{align}
\mathcal{L}_\text{cl} & = \frac{1}{n} \sum_{i=1}^{n} \mathcal{L}_\text{InfoNCE}(i) + \frac{1}{n} \sum_{i=1}^{n}\mathcal{L}_\text{InfoNCE}'.
\end{align}

\subsection{Pre-Training Objective}
\label{sec:objective}
We combine the two pre-training methods discussed above to define the overall objective as: 
\begin{align}
    \mathcal{L} = \mathcal{L}_\text{mask} + \alpha \gamma \mathcal{L}_\text{cl},
\end{align}
where $\alpha$ is a hyperparameter controlling the weight of the contrastive loss and multiplying the temperature $\gamma$ will smooth training \cite{grill2020byol, chen2021mocov3}.
The inputs for both tasks are shared in mini-batches with the \ourcl loss using the same block-masked inputs necessary for the \ourmask task. 
We highlight that the masked tokens for the denoising task are the only noise introduced in the contrastive learning, and that no other data augmentation is applied to raw pixels, in contrast to previous vision contrastive learning methods in which data-augmentation was paramount to the final performances of the model. 
This phenomenon is empirically studied in \sref{sec:cl_analysis}.

%% file: 4_model.tex
\subsection{Modeling}
\label{sec:backbone}
The model architecture follows the standard transformer architecture in its post-layer-norm variant~\cite{vaswani2017attention, devlin2019bert} with two more recent additions: divided temporal-spatial attention~\cite{bertasius2021space}, and sparse spatial attention~\cite{child2019generating}.
The model embedding layer maps the discrete tokens $\{x_{t,i,j}\}$ of a quantized input video (see \sref{sec:mask_task}) into dense vectors and sum them with positional embeddings. The backbone transformer model then outputs corresponding features $\{h_{t,i,j}\}$. We  append an additional \texttt{[CLS]} token to each input sequence following~\cite{devlin2019bert} and use its output feature $h_\text{cls}$ as a representation for the whole video.
For pre-training, we use two heads: a 2-layer MLP after each token outputs $\{h_{t,i,j}\}$ for the \ourmask task following BERT~\cite{devlin2019bert}, and a 3-layer MLP after the CLS output $h_\text{cls}$ for the \ourcl task following SimCLR~\cite{chen2020simple}.
For fine-tuning on classification tasks, we remove the pre-training heads and add a fully-connected layer to the \texttt{[CLS]} output $h_\text{cls}$.
Please see \sref{sec:appen_model} for a detailed model description.

%% file: 5_setup.tex
\section{Experiments and Results}

\begin{figure}[t]
\vskip 0.1in
\begin{center}
\includegraphics[
                width=0.95\columnwidth,
                ]{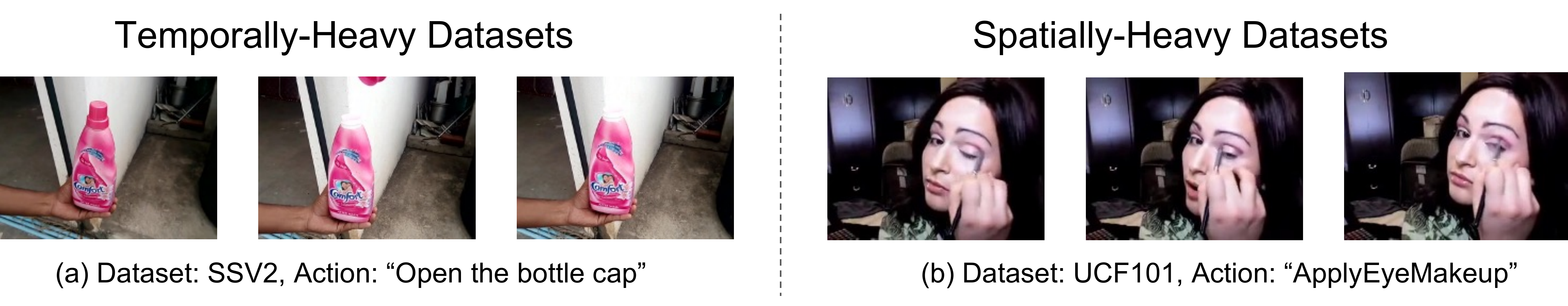}
                \vspace{-6pt}
\caption{
Example videos from \textit{temporally-heavy} (\textit{left}) and \textit{spatially-heavy} (\textit{right}) datasets. For the spatially-heavy video example, we can easily recognize the action `apply eye makeup' from any of its frames. However, in order to recognize the action `open the bottle cap' from the temporally-heavy data example, we need to consider the relations between multiple frames.  
}
\label{fig:temporal_vs_spatial_small}
\end{center}
\vspace{-10pt}
\end{figure}

\subsection{Datasets}
\label{sec:datasets}
For \textbf{pre-training}, we use the HowTo100M dataset~\cite{miech19howto100m}.
This dataset is constructed by searching YouTube videos with a list of text queries, it is significantly larger and more diverse than human-annotated datasets such as Kinetics 400~\cite{carreira2019short}. 
HowTo100M has 1.2M uncurated videos, with an average duration of 6.5 minutes. 
We only use videos and do not use other signals such as ASR captions in this dataset.
For \textbf{downstream} evaluation, we experiment with several action classification datasets: \ucf~\cite{Soomro2012UCF101AD}, \hmdb~\cite{Kuehne2011HMDBAL}, \ksmall~\cite{carreira2017quo}, \ssv~\cite{goyal2017something}, and \diving~\cite{li2018resound}. 
It is important to note that in many cases, actions in \ucf, \hmdb, and \ksmall can be recognized from a single frame of the video, thus these datasets are `\emph{spatially-heavy}'.
As a consequence, image-level methods~\cite{bertasius2021space, radford2021learning} show competitive results without modeling the temporal interactions inside the videos.
To test the video model's ability beyond recognizing static images, we lay our focus on `\emph{temporally-heavy}' datasets (\ssv and \diving), in which action recognition from a single frame is more difficult. For example, it is almost impossible to distinguish two \ssv classes \textit{open something} and \textit{close something} without reasoning across frames, and the same for different diving classes in \diving. 
We show examples of temporally-heavy and spatially-heavy datasets in \fref{fig:temporal_vs_spatial_small}. More examples are in \sref{sec:append_dataset_visualization}.
Additional dataset details and statistics are presented in \sref{sec:appen_dataset}.

\subsection{Experimental Setup}
Our model shapes follow BERT$_\text{LARGE}$ with 24 layers and hidden size 1024, but with halved attention head size and MLP intermediate size as in~\cite{child2019generating}.
For \textbf{pre-training}, we train the model for $100$ epochs on the HowTo100M dataset with frames sampled at 2 FPS.
We create the training inputs by sampling two clips from each video as described in Sec.~\ref{sec:contrastive}.
To reduce computation cost, the first $90$ epochs are trained with a smaller input resolution (\#frames $T$=$5$ and frame size $S$=$128$) and we increase the spatial resolution ($T$=$5$, $S$=$256$) for the last $10$ epochs following~\cite{devlin2019bert}. 
Positional embeddings are interpolated as in \cite{dosovitskiy2020image} when input resolution changes.
Importantly, our pre-training scheme does not involve spatial augmentations: all frames are resized and centered cropped without random flipping, color distortion, etc. 
We use a batch size of 1024 in pre-training.
The number of negative clips used for \ourcl is 255 for the first 90 epochs and 127 for the last 10 epochs.
The number of negative pairs used in our ablation analyses is kept constant at 127.\footnote{During pre-training, we always accumulate the gradient to a batch size of 1024 before updating the weights but use different numbers of negative examples. We analyze this effect in \sref{sec:cl_analysis}.}
For \textbf{fine-tuning}, we use more input frames ($T$=$10$ and $S$=$256$), and batch size 128.
We sample frames at 2 FPS for datasets with longer videos (i.e., \ucf and \ksmall), and sample 4 FPS for datasets with shorter videos (i.e., \hmdb, \ssv, \diving).
During inference, we follow~\cite{feichtenhofer2019slowfast, feichtenhofer2021large} to use 3 spatial crops and 10 temporal crops (in total 30 crops), and average their prediction scores as the final score.\footnote{As in~\cite{bertasius2021space, arnab2021vivit}, we observe that the performance is saturated at 4$\sim$5 temporal crops for our model. For SSV2, we use $10$ crops following~\cite{feichtenhofer2021large}. Details are in \sref{sec:appen_experiment}}
All models are trained with AdamW~\cite{loshchilov2018decoupled} optimizer with linear warm-up and linear learning rate decay.
We observe similar pre-training instability as reported in \cite{chen2020generative,chen2021mocov3} and follow their practice to  sequentially choose learning rate at 1e-3, 5e-4, 3e-4, ..., until convergence.
More details for experiments are listed in \sref{sec:appen_experiment}.

%% file: 6_result.tex
\begin{table}
\caption{
\textbf{Comparison with state-of-the-art.}
Our model outperforms previous works on \ssv and \diving dataset while showing competitive results on other datasets. 
 UCF101 and HMDB51 are average over three train-val splits.
 V,A,T refer to Visual, Audio, and Text modalities, respectively.
 AS and HTM are Audio Set\cite{gemmeke2017audio} and HowTo100M~\cite{miech19howto100m}.
 }
\resizebox{\textwidth}{!}{%
\begin{tabular}{@{}llllllll@{}}
\toprule
\multirow{2}[2]{*}{Method} & \multirow{2}[2]{*}{Modality} & \multirow{2}[2]{*}{Pre-Train Dataset} & \multicolumn{2}{c}{Temporally-Heavy}    & \multicolumn{3}{c}{Spatially-Heavy} \\
\cmidrule(lr){4-5} \cmidrule(lr){6-8}
                                & & & SSV2~\cite{goyal2017something}            & \diving~\cite{li2018resound}           & UCF101~\cite{soomro2012ucf101} & HMDB51~\cite{kuehne2011hmdb} & K400~\cite{carreira2017quo}           \\ \midrule
                                
Previous SotA                   & V & -                & 65.4~\cite{arnab2021vivit}            & 81.0~\cite{bertasius2021space}              & 98.7~\cite{kalfaoglu2020late}   & 85.1~\cite{kalfaoglu2020late}   & 84.8~\cite{arnab2021vivit}          \\ 
\;\;\;w/o Temporal    & V & -                & 36.6~\cite{bertasius2021space}            & -                 & 92.0~\cite{radford2021learning}   & -      & 77.6~\cite{bertasius2021space}              \\ \midrule
\multicolumn{7}{l}{\footnotesize{\textit{Self-supervised Pre-Training}}} \\
K400 Self-Sup.          & V & Kinetics-400~\cite{carreira2017quo}     & 55.8~\cite{feichtenhofer2021large}            & -                 & 96.3~\cite{feichtenhofer2021large}   &  75.0~\cite{feichtenhofer2021large}      & -       \\
MIL-NCE~\cite{miech2020end}                         & V+T& HTM~\cite{miech19howto100m}             & -               & -                 & 91.3   & 61.0   & -                 \\
MMV~\cite{alayrac2020self}                       & V+A+T& AS~\cite{gemmeke2017audio}+HTM~\cite{miech19howto100m}             & -               & -                 & 95.2   & 75.0   & -                 \\
MoCo~\cite{feichtenhofer2021large}           & V & IG-Uncurated~\cite{ghadiyaram2019large}     & 53.2            & -                 & 92.9   & -      & - \\
\textbf{\modelname}            &V  & HTM~\cite{miech19howto100m}             &68.1            & 85.5             & 92.7   & 65.9    & 77.4               \\ \midrule
\multicolumn{7}{l}{\footnotesize{\textit{Supervised Pre-Training}}} \\
K400 Sup.                 & V &  \ksmall~\cite{carreira2017quo}          & 63.1~\cite{feichtenhofer2019slowfast}                &                -   & 96.8~\cite{tran2018closer}   & 82.5~\cite{wang2019hallucinating}   &  81.5~\cite{kondratyuk2021movinets}         \\ 
TimeSformer~\cite{bertasius2021space}                     &V& ImageNet-21K~\cite{russakovsky2015imagenet}     & 62.3            & 81.0              & -      & -      & 80.7              \\
ViViT~\cite{arnab2021vivit}     & V& ImageNet-21K~\cite{russakovsky2015imagenet}     & 65.4            & -                 & -      & -      & 80.6              \\
\bottomrule
\end{tabular}
}
\label{tab:results}
\end{table}

\subsection{Results}
\label{sec:results}
We compare our primary results with previous work in \tref{tab:results}.
We expand the results that are most related to our work: self-supervised training on uncurated videos and supervised pre-training with transformers.
For other results, we select the best-performing models to our knowledge and denote their reference in the table.

Our model \modelname sets the new state of the art on the two temporally-heavy datasets \ssv and \diving, where we achieve 2.7\% and 4.5\% absolute improvement, respectively, over previous best models among all self-supervised and supervised pre-trained methods.
This is especially surprising considering the two previous SotA models ViViT~\cite{arnab2021vivit} and TimeSformer~\cite{bertasius2021space} both use large-scale supervised pre-training, and ViViT also uses various regularization techniques (e.g., stochastic depth~\cite{huang2016deep}, random augment~\cite{cubuk2020randaugment} and mixup~\cite{zhang2018mixup}).
\modelname also achieves competitive results on other three spatially-heavy datasets: \ucf, \hmdb, and \ksmall.
As discussed in \sref{sec:datasets}, recognizing actions in \ssv and \diving require a strong temporal reasoning ability, while in the other datasets, spatial understanding is dominant.
Some relatively low results of our \modelname (e.g., K400) are thus possibly due to the VQ-VAE spatial information loss.
To illustrate this, we show a comparison between the SotA models~\footnote{
Some SotA models are pre-trained with extremely large (weakly-)supervised datasets, e.g., IG65M~
\cite{ghadiyaram2019large} in \cite{kalfaoglu2020late} and JFT-300M~\cite{sun2017revisiting} in \cite{arnab2021vivit}.
} with temporal modeling in the first row and the ones without in the second row (`w/o Temporal') of \tref{tab:results}.
Note the gaps between these two types of models are significantly larger for temporally-heavy datasets (\ssv) than spatially-heavy datasets (\ucf, \ksmall), demonstrating the importance of temporal modeling for temporally-heavy datasets.
We also show the methods pre-trained on \htm that take other modalities to help video learning thus beyond the scope of visual self-supervised learning.

Previous self-supervised pre-training such as MoCo~\cite{feichtenhofer2021large} are good at global understanding, but the pre-training schema does not consider the internal interactions inside videos (especially for the temporal dimensions).
As a result, it could reach or even outperform the supervised alternatives on \ucf.
However, it shows lower results on \ssv compared to the transformer models~\cite{bertasius2021space, arnab2021vivit} (although with different backbone models) that warm up from image-pre-trained models and learn the temporal interactions directly from the downstream tasks.

%% file: 7_analysis.tex
\section{Analysis}
\label{sec:analysis}
We also analyze the model's scalability and the effectiveness of our pre-training methods.
To save computation, for all analyses, we use a smaller model (6-layer transformer with hidden dimension 512) and smaller input resolution (5 input frames with spatial size 128, i.e., $T$=$5$, $S$=$128$) throughout this section, unless otherwise stated. 
We also perform pre-training with fewer epochs (i.e., 10).
For downstream tasks, we use the same input resolution as pre-training (i.e., $T$=$5$, $S$=$128$), and we use $2$ temporal crops for inference.
All results are reported on the train-val split 1 if applicable.

\begin{table}
\centering
\caption{ 
Impact of \textbf{model size}.
`Params' is the number of parameters.
`Speed' is the normalized pre-training speed measured by \#videos/second on one V100 GPU. 
`Mask-Accu.' and `CL-Loss' are \ourmask accuracy and \ourcl loss to indicate the pre-training performance.
`UCF101' is the fine-tuning accuracy on UCF101 dataset.
First line is defautly used in analysis and the configuration producing final results are underlined.
}
\resizebox{0.67\textwidth}{!}{%
\begin{tabular}{@{}cc|cc|ccc@{}}
\toprule
Layers  & Dim   & Params       & Speed & Mask-Accu.$\uparrow$   & CL-Loss $\downarrow$  &  UCF101$\uparrow$   \\ \midrule
6       & 512   & 29.4M         &  32.0          & 17.2         & 1.06      & 69.4                             \\ \midrule
6       & 768   & 63.0M         &  21.0          & 17.7         & 1.03      & 75.0                             \\
12      & 512   & 54.7M         &  18.1          & 17.9         & 1.02      & 76.6                             \\
12      & 768   & 119.7M        &  11.2          & 18.4         & 1.00      & 78.1                          \\
\underline{24}      & \underline{1024} & 210.1M        &  5.0           & 18.7         & 0.98      & 78.5                                 \\
\bottomrule
\end{tabular}%
 }
\label{tab:scale}
\end{table}

\subsection{Scalability} 
\label{sec:model_scale_analysis}

\textbf{Model.}
In \tref{tab:scale}, we illustrate the scalability of our method with different model sizes (i.e., number of layers and hidden dimensions).
Larger models have more parameters (`Params') and higher computational cost (measured by the normalized pre-training `Speed').
To evaluate the pre-training tasks performance, we provide both pre-training metrics (\ourmask accuracy denoted by `Mask-Accu.', and \ourcl loss denoted by `CL-Loss') and \ucf downstream fine-tuning results.
As the size of the model grows, the fine-tuning results show consistent improvement with the pre-training metrics. 
Note that for the last row in \tref{tab:scale}, we halve the attention head and MLP intermediate dimensions.

\begin{table}
\caption{Impact of \textbf{input resolutions $T$ and $S$}. 
`Mask-Accu.' and `CL-Loss' are the pre-training metrics. 
`UCF101' indicates the \ucf fine-tuning results with the pre-training resolution.
`UCF101-Full-Reso.' indicates the full-resolution fine-tuning with $T$=10 and $S$=256.}
\resizebox{\textwidth}{!}{%
\begin{tabular}{cc|cc|cccc}
\toprule
\#frames $T$            & Frame Size $S$           & Params & Pre-train Speed & Mask-Accu.$\uparrow$ & CL-Loss$\downarrow$ & UCF101$\uparrow$ & UCF101-Full-Reso.$\uparrow$ \\ \midrule
5                 & 128                  & 29.4M  & 32.0            & 17.2      & 1.06    & 69.4  & 73.8   \\
10                & 128                  & 29.4M  & 16.5            & 17.2      & 0.96    & 74.2  & 74.6   \\
5                 & 256                  & 29.4M  & 8.4             & 10.8      & 0.93    & 72.9  & 75.7   \\
10                & 256                  & 29.4M  & 4.4             & 10.6      & 0.85    & 78.1  & 78.1   \\
\bottomrule
\end{tabular}%
}
\label{tab:input_resolution}
\end{table}

\textbf{Input Resolution.}
In \tref{tab:input_resolution}, we show model scalability over input resolution (i.e., \#frames $T$ and frame size $S$).
With the same frame size $S$, longer clips perform better than shorter clips (e.g., $T$=10, $S$=128 is better than $T$=5, $S$=128).
With the same number of input frames $T$, larger frame size improves the performance (e.g., $T$=10, $S$=256 is better than $T$=10, $S$=128).
For each pre-training resolution, we also try to fine-tune under a full-resolution with $T$=10, $S$=256 (denoted as `UCF101-Full-Reso.'). 
As in pre-training, fine-tuning with larger resolution generally improves the results.
Although longer and smaller clips ($T$=10, $S$=128) show better results than shorter and larger clips ($T$=5, $S$=256) when using the same pre-training and fine-tuning resolutions, they show different trends with the full-resolution fine-tuning.
Increasing frame size during fine-tuning (the second block in \tref{tab:input_resolution}) only improves the \ucf result by $0.4$, while increasing the clip length (the third block) improves the \ucf result by 3.8.
These results call for a need of pre-training with large spatial size, and we follow this practice in our large-scale experiments as in \sref{sec:results}.

\subsection{Pre-Training Methods}
We analyze the key designs of our two pre-training tasks.
When analyzing the \ourmask task in \sref{sec:mask_analysis}, we exclude the \ourcl loss (by setting loss ratio $\alpha\mbox{=}0$) to preclude potential side effects.
However, we still use masked prediction loss when assessing the \ourcl task in \sref{sec:cl_analysis} as we observe very low performance with only \ourcl objective.

\subsubsection{The Impact of Pre-Training}
\label{sec:pretrain_analysis}

\begin{wraptable}[10]{r}{0.45\textwidth}
\setlength{\tabcolsep}{0.15em}
\centering
\vspace{-25pt}
\caption{
Impact of \textbf{pre-training tasks}. `MP'=Mask-then-Predict, `CL'=Contrastive Learning task.
}
\resizebox{0.45\textwidth}{!}{%
\begin{tabular}{@{}ccccccc@{}}
\toprule
\multirow{2}[2]{*}{MP} & \multirow{2}[2]{*}{CL} & \multicolumn{2}{c}{Temporally-Heavy}    & \multicolumn{3}{c}{Spatially-Heavy} \\
\cmidrule(lr){3-4} \cmidrule(lr){5-7}
                            & & SSV2 & Diving48 & UCF101 & HMDB51 & K400 \\ \midrule
\xmark         & \xmark             & 1.2  & 10.0     & 41.3   &19.0    & 41.0         \\ 
\xmark         & \cmark     & 32.5 & 26.3     & 57.1   &30.7    & 47.0     \\
\cmark & \xmark      & 41.4 & 37.2     & 68.3   &35.3    & 53.7     \\
\cmark & \cmark     & 41.1 & 37.5     & 69.4   &37.8    & 54.5         \\
\bottomrule
\end{tabular}%
}
\label{tab:non_pretraining}
\end{wraptable}

We first compare different pre-training tasks and the non-pre-training results.
As shown in \tref{tab:non_pretraining}, \ourmask is good at temporally-heavy datasets (\ssv, \diving) while \ourcl improves the spatially-heavy datasets.
We also compare with the non-pre-training results (the first row of \tref{tab:non_pretraining}) and observe that both tasks significantly improve the results.
We notice that these non-pre-training results are lower than previous from-scratch models, which might be caused by the difficulty in training video transformers~\cite{bertasius2021space, arnab2021vivit} and the information loss in our input quantization process~\cite{ramesh2021zero}.

\begin{figure}[t]
\vskip 0.1in
\begin{center}
\includegraphics[
                width=0.95\columnwidth,
                ]{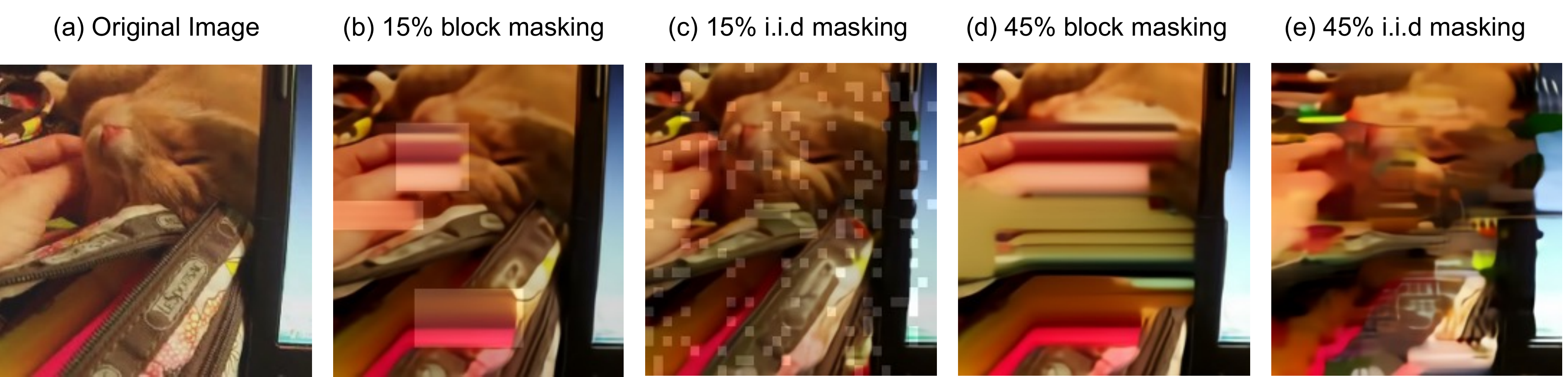}
                \vspace{-4pt}
\caption{
Nearest-neighbour reconstruction of \textit{block masking} and \textit{i.i.d masking}. Figure (a) is the original image, and other figures are reconstructions under different masking strategies.
}
\label{fig:block_vs_iid_small}
\end{center}
\vspace{-1pt}
\end{figure}

\subsubsection{Mask-then-Predict}
\label{sec:mask_analysis}

\begin{table}[t]
\setlength{\tabcolsep}{0.2em}
\begin{minipage}[t]{.48\textwidth}
 \centering
\caption{ Impact of \textbf{masking strategies}. Models are pre-trained with only mask-then-predict.  }
\scalebox{0.9}{
\begin{tabular}{@{}cc|cl@{}}
\toprule
Strategy & Frame-Size $S$     & Mask-Accu.$\uparrow$       & UCF101 $\uparrow$  \\ \midrule
block            & 128            &  17.6                      & 68.3      \\
i.i.d.           & 128            &  24.3                      & 63.5 \textcolor{Salmon}{(-4.8)}          \\ \midrule
block            & 256            &  11.2                      & 69.5      \\ 
i.i.d.           & 256            &  19.5                      & 61.4 \textcolor{Salmon}{(-8.1)}           \\\bottomrule
\end{tabular}%
}
\label{tab:masking_strategy}
\end{minipage}
\hfill
\begin{minipage}[t]{.48\textwidth}
\centering
\caption{ Impact of \textbf{masking ratios}. Models are pre-trained with only mask-then-predict.  
Default setup is underlined.  }
\scalebox{0.9}{
\begin{tabular}{@{}c|cc|cc@{}}
\toprule
Strategy & \#Blocks & Ratio & Mask-Accu.$\uparrow$                  & UCF101  $\uparrow$        \\ \midrule
block            & 4              & 11.9\%        &  17.9                       & 66.8            \\
\underline{block}            & \underline{5}              & \underline{14.5\%}        &  17.6                       & 68.3           \\
block            & 6              & 17.0\%        &  17.3                       & 67.3            \\ 
\bottomrule
\end{tabular}%
}
\label{tab:masking_ratio}
\end{minipage}
\vspace{-1pt}
\end{table}

\textbf{Block Masking versus I.I.D. Masking.} 
We first compare our proposed block masking strategy and the uniform i.i.d. masking strategy (discussed in \sref{sec:mask_task} and illustrated in \fref{fig:block_mask_fig}).
As shown in \tref{tab:masking_strategy}, although the i.i.d. masking achieves higher pre-training mask-token-prediction accuracy (`Mask-Accu.'), it shows lower downstream results (`UCF101') than block masking.
The higher mask accuracy is possibly due to the easier i.i.d. \ourmask task.
To illustrate this, we show that simply copy-pasting from nearest neighbours yields reasonable reconstruction results in \fref{fig:block_vs_iid_small}. 
More discussions are provided in \sref{sec:append_mask_visualization}.
The existence of such a trivial solution potentially prevents the model from learning useful video representations for downstream tasks.
Meanwhile, we also find that the model with larger input frame size $256$ benefits more from the block masking strategy, because the adjacent tokens are closer in the original 2D image for these larger frames.
Hence, the spatial locality is amplified.

\textbf{Masking Ratio.}
In \tref{tab:masking_ratio}, we study the impact of masking ratio, by varying the number of masked blocks for block masking. 
Empirically, the result differences among different masking ratios are marginal and the original BERT's $15$\% masking ratio (with roughly $5$ masking blocks) works slightly better.
Thus we always select the number of mask blocks whose induced masking ratio is closest to $15$\%.
The detailed choices of number of masking blocks are listed in \sref{sec:appen_masking_blocks}.

\begin{figure}[t]
\vskip 0.1in
\begin{center}
\includegraphics[
                width=0.95\columnwidth,
                ]{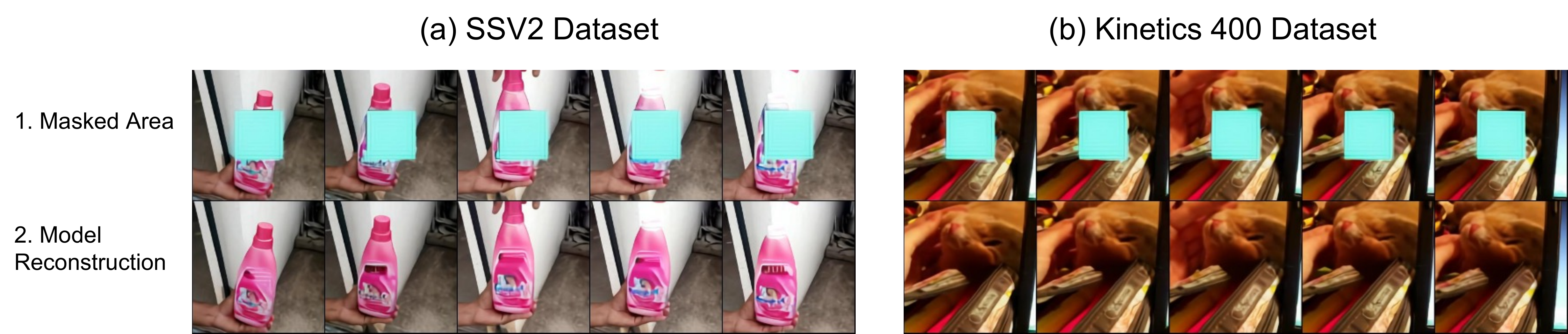}
                \vspace{-4pt}
\caption{
Model reconstruction results for SSV2 and Kinetics-400 datasets.
}
\label{fig:reconstruction_small}
\end{center}
\vspace{-2pt}
\end{figure}

\textbf{Reconstruction Visualization.}  
In \fref{fig:reconstruction_small}, we show our model reconstruction results from block masking.
The centering block in the image is masked, and our model can reconstruct the image patch with the spatio-temporal context.
Details are in \sref{sec:append_recon_visualization}.

\subsubsection{Contrastive Learning}

\label{sec:cl_analysis}

\begin{table}[t]
\setlength{\tabcolsep}{0.2em}
\begin{minipage}[t]{.48\textwidth}
 \centering
\caption{
Impact of \textbf{maximum sampling distance} $d_\text{max}$ (seconds) between two positive clips.
}
\scalebox{0.9}{
\begin{tabular}{l|ccl}
\toprule
$d_\text{max}$       & Mask-Accu.$\uparrow$ & CL-Loss$\downarrow$ & UCF101$\uparrow$ \\ \midrule
\underline{$\infty$}   & 17.2      & 1.06    & 69.4  \\
30         & 17.3      & 0.77    & 69.0 \textcolor{Salmon}{(-0.4)}  \\
10         & 17.4      & 0.61    & 68.3 \textcolor{Salmon}{(-1.1)} \\
0          & 17.5      & 0.41    & 66.7 \textcolor{Salmon}{(-2.7)} \\
\bottomrule
\end{tabular}%
}
\label{tab:cl_distribution}
\end{minipage}
\hfill
\begin{minipage}[t]{.48\textwidth}
 \centering
\caption{
Impact of \textbf{\#negative samples}. 
}
\scalebox{0.9}{
\begin{tabular}{l|ccc}
\toprule
\#neg-samples        & Mask-Accu.$\uparrow$ & CL-Loss$\downarrow$ & UCF101$\uparrow$ \\ \midrule
\underline{128} - 1            & 17.2      & 1.06    & 69.4  \\
256 - 1               & 17.1      & 1.30    & 69.2  \\
512 - 1               & 17.2      & 1.56    & 70.4    \\
1024 - 1               & 17.0      & 1.86    & 69.8  \\
\bottomrule
\multicolumn{4}{c}{} \\
\end{tabular}%
}
\label{tab:batch_size}
\end{minipage}
\vspace{-15pt}
\end{table}

\textbf{Positive Sampling Distance.}
As illustrated in \sref{sec:contrastive} and \fref{fig:block_mask_fig}.(b), we uniformly sample positive clip pairs across the whole video without any distance restriction.
To analyze the effect of such a sampling strategy, 
We perform a set of experiments by varying the maximum sampling distance $d_{max}$ (in seconds) between two positive clips.
The results are shown in \tref{tab:cl_distribution}.
$d_{max}$=$\infty$ denotes our default setup without any distance restriction. 
$d_{max}$=$0$ samples two same clips, and $d_{max}$=$10$ samples two positive clips with a maximum distance of 10 seconds.
Although previous contrastive learning methods~\cite{qian2020spatiotemporal, feichtenhofer2021large} favor the sampling of temporal positives within a shorter range (e.g., maximum $36$ seconds for uncurated videos in~\cite{feichtenhofer2021large}),
we observe a performance gain when using larger distance.
We also want to emphasize that the results with $d_{max}$=10 and $d_{max}$=0 are not better than the model pre-trained with only \ourmask (\ucf accuracy 68.3), which suggests that short-range contrastive learning does not improve upon our \ourmask task.
This is potentially because our \ourmask already gives the model the ability to model local interactions, thus \ourcl objective can only be useful when it focuses on longer-range interactions.

\textbf{Number of Negative Samples.}
Previous constrastive learning methods \cite{chen2020simple,chen2021mocov3,feichtenhofer2021large} benefit from more negative samples.
In this section, we show that the number of negative samples has less impact on our method when \ourmask task is added.
As shown in \tref{tab:batch_size}, we experiment with different \ourcl sample sizes (i.e., $n$ in \sref{sec:contrastive} which is 1 + number of negative samples) and always accumulate the gradients to $1024$ samples before updating the parameters.
Although increasing sample size makes the \ourcl task harder (reflected by `CL-Loss'), it does not show clear evidence of improving \ucf downstream performance.

\begin{wraptable}[11]{r}{0.45\textwidth}
\setlength{\tabcolsep}{0.15em}
\centering
\vspace{-10pt}
\caption{
Impact of \textbf{mask augmentation in \ourcl}. 
`MP'=Mask-then-Predict.
`CL-Mask'=Use input mask in CL.
Default setup is underlined.
}
\resizebox{0.45\textwidth}{!}{%
\begin{tabular}{cc|ccc}
\toprule
MP    &CL-Mask       & Mask-Accu.$\uparrow$ & CL-Loss$\downarrow$ & UCF101$\uparrow$ \\ \midrule
\xmark           & \xmark            & -     & 1.07    & 57.1  \\
\xmark           &\cmark               & -     & 1.08    & 55.5  \\
\cmark           &\xmark            & 17.2  & 1.04    & 67.4  \\
\underline{\cmark}               &\underline{\cmark}               & 17.2  & 1.06    & 69.4  \\
\bottomrule
\end{tabular}%
}
\label{tab:mask_in_cl}
\end{wraptable}

\textbf{Input Masking as Augmentation.}
Most self-supervised visual contrastive learning methods~\cite{chen2020improved, chen2020simple, grill2020byol,feichtenhofer2021large, qian2020spatiotemporal} suffer from a large drop when removing strong spatial augmentations.
In contrast, our pre-training does not use any spatial augmentations on raw frames, such as flipping and random cropping.
However, as we tie the input between \ourmask and \ourcl to reduce computation cost, the random masking noise is naturally introduced.
We here investigate its impact in \tref{tab:mask_in_cl}.
When pre-trained jointly with \ourmask, adding mask noise improves \ucf accuracy by +2.0;
however, when pre-trained without it, adding mask noise hurts the performance (-1.6).
We hypothesize that this is due to the large input mismatches between pre-training and fine-tuning when \ourmask objective is not applied.
Noisy masking creates `holes' in the input token maps during pre-training, while for fine-tuning the input token maps are intact. 
When \ourmask task is applied, it guides the model to fill these holes, thus reducing this mismatch and allowing the \ourcl task to benefit from noisy masking as a type of regularization.
In contrast, this input mismatch becomes dominant when only using the \ourcl objective.

%% file: appendix.tex
\appendix

\section{Model Architecture}
\label{sec:appen_model}
As described in \sref{sec:backbone}, we use a transformer model on top of the discrete video tokens generated by VQ-VAE. Since transformers have different variants, we here show details of our architecture for clarity.
The design of our method largely follows the practice in BERT~\cite{devlin2019bert}, TimeSformer~\cite{bertasius2021space}, Sparser Transformer~\cite{child2019generating}, ViViT~\cite{arnab2021vivit}, and MoCoV3~\cite{chen2021mocov3}.

\subsection{Backbone Model}

\paragraph{Embedding.} 
Given the video frames $\{f_t \in \mathbb{R}^{H \times W} \mid t \in [T]\}$, we first use VQ-VAE~\cite{van2017neural,ramesh2021zero} to discrete them into video tokens $\{x_{t,i,j} \in [V] \mid t\in [\hat{T}], i \in [\hat{H}], j \in [\hat {W}] \}$. 
We next use an embedding layer ($\mathit{embedding}$) that to map these discrete tokens to continuous vectors. 
Since transformer layers are permutation-invariant, we follow~\cite{dosovitskiy2020image,devlin2019bert} to add positional information into the input. 
The positional embedding ($\mathit{pos}$) is factorized as a sum of the temporal embedding $\mathit{pos}^\textsc{t}$, the height embedding $\mathit{pos}^\textsc{h}$, and the width embedding $\mathit{pos}^\textsc{w}$. 
This factorization reduces the number of trainable parameters to encode positional information, which empirically shows a slightly better result.
Finally, a Layer-Normalization~\cite{ba2016layer} layer is added to get the initial hidden outputs $h^0_{t,i,j}$:
\begin{align}
h^0_{t,i,j} &= \mathrm{LayerNorm}(\mathit{embedding}(x_{t,i,j}) + \mathit{pos} (t, i, j)), \\
\mathit{pos}(t, i, j) & = \mathit{pos}^\textsc{t}(t) +  \mathit{pos}^\textsc{h}(i) + 
\mathit{pos}^\textsc{w}(j),
\end{align}
where we use the superscript 0 to denote that it is the hidden outputs before the first transformer layer.

\paragraph{Attention Blocks.}
Before introducing the detailed model architecture, we first describe the basic building components: the attention block.
An attention block is built based on the attention operator (i.e., `$\mathrm{Attn}$') with a residual connection.
The attention operator takes a single query vector $x$ and its context $\{y_i\}$ as input.
It first computes the attention score between $x$ and each context vector $y_i$, then the attention scores are normalized by the $\mathrm{softmax}$.
Lastly, the output is a weighted-sum over all the context vectors (transferred by a `value' matrix $W_\text{value}$):
\begin{align}
 \mathrm{Attn}(x, \{y_i\}) & = \sum_i \mathrm{softmax}_i\{ (W_\text{query} \, x)^\top \,W_\text{key} \, y_i\} W_\text{value} \, y_i.
\end{align}
To compose the attention block from the previous attention operator, the residual connection and layer normalization (i.e., `$\mathrm{LayerNorm}$') are added.
We follow the original transformer model~\cite{vaswani2017attention} that uses a post-layer-norm layout:
\begin{align}
 \mathrm{AttnBlock} (x, \{y_i \}) &= \mathrm{LayerNorm}( x + W_\text{out}\, \mathrm{Attn}(x, \{y_i\})).
\end{align}
In order to reduce computational cost and memory, we also adapt the attention block suggested in Sparse Transformer~\cite{child2019generating} that takes two sets of context vectors $\{y_i\}$ and $\{z_j\}$ as input.
This special attention block computes attention for the two context-vector sets separately and concatenates their output together.
In our case, suppose $\{y_i\}$ and $\{z_j\}$ are the rows and columns of a square matrix, then it reduces the computation cost of calculating attention scores from $\Theta(n^4)$ to $\Theta(n^2)$, where $n$ is the number of rows/columns:
\begin{align}
 \mathrm{AttnBlock} (x, \{y_i \}, \{z_j\}) &= \mathrm{LayerNorm}( x + W_\text{out}\, [\mathrm{Attn}(x, \{y_i\}), \mathrm{Attn}(x, \{z_j\})]) 
\end{align}
\paragraph{Spatio-Temporal Transformer Layer.} 
The spatio-temporal transformer layer is composed with the previously-introduced attention blocks and an additional MLP block.
The $l$-th layer takes the output of the previous layer $\{h^{l-1}_{t, i, j}\}$ as input and outputs the hidden states $\{h^l_{t, i, j}\}$.
We separate the attention into two attention blocks: the temporal attention block $\mathrm{AttnBlock_\textsc{time}}$ and the spatial attention block $\mathrm{AttnBlock_\textsc{space}}$.
Without loss of generality, we will use $g^\textsc{time}$ and $g^\textsc{space}$ to denote the intermediate results from temporal and spatial attention blocks, respectively.
First, the temporal attention block attends to the tokens at the same spatial location but in different frames (i.e., at different timesteps): $\{h^{l-1}_{t, i, j} \mid t \in [T] \}$.
Next, the spatial attention block attends to the tokens in the same frame: $\{g^\textsc{t}_{t, i, j} \mid (i, j) \in [\hat{H}] \times [\hat{W}] \}$.
To reduce the computational cost, we incorporate the sparse attention block~\cite{child2019generating} (detailed in the previous paragraph) that factorizes the attention over height and width: $\{g^\textsc{t}_{t, i, j} \mid i \in [\hat{H}] \}$, $\{g^\textsc{t}_{t, i, j} \mid j \in [\hat{W}] \}$.
The MLP block has two fully-connected layers with GeLU~\cite{hendrycks2016gaussian} activation in the middle.
Overall, the formula of one spatio-temporal transformer layer is:
\begin{align}
    g^\textsc{time}_{t, i, j} &= \mathrm{AttnBlock_\textsc{time}} (h^{l-1}_{t, i, j}, \{h^{l-1}_{t, i, j} \mid t \in [T] \}) \\
    g^\textsc{space}_{t, i, j} &= \mathrm{AttnBlock_\textsc{space}} (g^\textsc{t}_{t, i, j}, \{g^\textsc{time}_{t, i, j} \mid i \in [\hat{H}] \}, \{g^\textsc{time}_{t, i, j} \mid j \in [\hat{W}] \}) \\ 
    h^{l}_{t, i, j} &=  \mathrm{LayerNorm}(g^\textsc{space}_{t, i, j} + \mathrm{MLP}(g^\textsc{s}_{t, i, j})) 
\end{align}
\paragraph{\texttt{[CLS]} Token.} 
Following the practice in BERT~\cite{devlin2019bert} design, we add a special \texttt{[CLS]} (abbreviation of `classification') token and take its output as the representation of the whole sequence.
We follow TimeSformer~\cite{bertasius2021space} to compute the its output: the \texttt{[CLS]} token attends over the context separately and then the outputs are averaged. We take the temporal attention layer as an example.
Suppose $h_\text{cls}^{l-1}$ is the \texttt{[CLS]} feature vector output by layer $l-1$, then the temporal attention layer do the following computation:
\begin{align}
    g^\textsc{time}_\text{cls} = \frac{1}{\hat{H}}\frac{1}{\hat{W}}\sum_{i}\sum_{j}\mathrm{AttnBlock_\textsc{time}} (h_\text{cls}^{l-1}, \{h^{l-1}_{t, i, j} \mid t \in [T] \}).
\end{align}
The other attention blocks process the \texttt{[CLS]} token similarly.

\subsection{Pre-Training and Fine-Tuning Heads}
\label{sec:appen_heads}
Pre-training or fine-tuning usually requires a few additional modules (i.e., heads) on top of the transformer layers that convert the output features to the desired probabilities or vectors.
We next describe the heads used in our pre-training and fine-tuning process.
\paragraph{Token Head for Mask-then-Predict.} 
We first define the prediction head over the tokens following BERT\cite{devlin2019bert}. 
It first processes the last-layer hidden outputs $h^L_{t, i, j}$ using a fully-connected layer (with GELU activation~\cite{hendrycks2016gaussian} and layer normalization~\cite{ba2016layer}):
\begin{align}
    u_{t,i,j} = \mathrm{LayerNorm} \left(\mathrm{GELU} (W_\text{token} (h^L_{t, i, j}) + b_\text{token})\right).
\end{align}
In our \ourmask method (\sref{sec:mask_task}), we will predict the masked tokens (i.e., the token before masking) from their context. 
We thus further convert this hidden vector into a distribution over the token vocabulary:
\begin{align}
    P_{t,i,j}(o_{t,i,j} = k) = \mathrm{softmax}_k \{ W_\text{word}\, u_{t,i,j} + b_\text{word} \}.
\end{align}
The weight $W_\text{word}$ is shared with input word embedding layer $\mathit{embedding}$ \cite{press2017using, devlin2019bert} while the bias $b_\text{word}$ is trained independently.

\paragraph{Contrastive Learning Head}
Next we discuss the pre-training heads for \ourcl. It is on top of the \texttt{[CLS]} hidden output $h_\textsc{cls}$.  We encode the hidden state with MLP.
We use batch normalization \cite{ioffe2015batch} inside the MLP head following the practice in \cite{chen2021mocov3}.
\begin{align}
    f_\textsc{cls} = \mathrm{MLP}_\textsc{cls} (h_\textsc{cls})
\end{align}
This $f_\textsc{cls}$ feature is used in computing the contrastive loss as in \sref{sec:contrastive}.

\paragraph{FC Layer for Fine-Tuning.}
When fine-tuning for action classification task, we add a fully-connected (FC) layer to the \texttt{[CLS]} output $h_\text{cls}$.
We initialize its weight and bias to zero.

\subsection{Special Tokens} 
Besides the $V$ token types introduced in the vocabulary of the VQ-VAE (see \sref{sec:mask_task}), we add several special tokens into the `vocabulary', namely a \texttt{[CLS]} token is introduced as a stub for the whole-video representation, 
a \texttt{[PAD]} token is used when the actual clip length is less than the model's expected input length. For the \ourmask task, we follow BERT~\cite{devlin2019bert} to replace the masked tokens with a specific \texttt{[MASK]} token.

\section{Pre-Training Details}
\label{sec:appen_pretrain}
\begin{table}
\centering
\caption{\textbf{Induced Masking ratio} w.r.t. to different input resolutions and \#masking blocks. The numbers of blocks/masking ratio for each resolution setting used in our experiments are shown in \textbf{bold}.}
\begin{tabular}{ccc|ccccc}
\toprule
\multicolumn{3}{c|}{Input Resolution} & \multicolumn{5}{c}{\#Masking Blocks}                          \\ \cline{4-8} 
Length          & Frame Size        & Token Map Size & 4    & 5             & 6             & 7             & 8    \\ \hline
5               & 128              & 16                & 11.9 & \textbf{14.5} & 17.0          & 19.4          & 21.7 \\
5               & 256               & 32                & 10.6 & 13.1          & \textbf{15.2} & 17.5          & 19.5 \\
10              & 128               & 16                & 10.4 & 12.8          & \textbf{15.0} & 17.1          & 19.2 \\
10              & 256               & 32                & 9.3  & 11.4          & 13.4          & \textbf{15.4} & 17.2 \\ \bottomrule
\end{tabular}%
\label{tab:appen_masking_blocks}
\end{table}

\subsection{Masking Blocks}

\label{sec:appen_masking_blocks}
As described in \sref{sec:mask_task}, we mask the tokens by blocks (a cube-shape set of tokens).
To avoid masking all the tokens in the clip, we control the maximum block length for the time domain, height, and width.
For spatial dimensions (i.e., height and width), the maximum length is half of the full length (e.g., the maximum block length will be 16 for a token map of length 32).
For temporal dimension (i.e., the clip length), the maximum length will be 2/3 (round up) of the full length so that it allows long-range modeling.
Under these constraints, we uniformly sample a fixed number of mask blocks and take their union to construct the final mask.
The number of blocks is decided by the induced masking ratio, which depends on the input resolutions.
In \tref{tab:appen_masking_blocks}, we show the induced masking ratio w.r.t. different input resolutions and \#masking blocks.
We take the VQ-VAE~\cite{van2017neural} provided in DALL-E~\cite{ramesh2021zero} that has a compression factor of 8, thus the length of the token map is always 1/8 of the frame size.
For each input resolution, we select the number of blocks (shown in bold in \tref{tab:appen_masking_blocks}) whose induced masking ratio is closet to $15$\% following BERT~\cite{devlin2019bert}.

\subsection{Contrastive Learning Loss}
For completeness, we list the two  losses used in contrastive learning here. The first loss for clip $c_i$ from $\text{video}_i$ is:
\begin{align}
\mathcal{L}_\text{InfoNCE}(i) &= - \log \frac{\exp{(f_i^\top f_i' / \gamma)}}{\sum_{k\neq i} \exp{(f_i^\top f_k / \gamma)} + \sum_{k}\exp{(f_i^\top f_k' / \gamma)}} 
\end{align}
The symmetric loss $\mathcal{L}'_\text{InfoNCE}(i)$ for feature of the other clip sample $c_i'$ from $\text{video}_i$ (and its feature $f_i'$) is:
\begin{align}
\mathcal{L}'_\text{InfoNCE}(i) = - \log \frac{\exp{(f_i'^\top f_i / \gamma)}}{\sum_{k} \exp{(f_i'^\top f_k / \gamma)} + \sum_{k \neq i}\exp{(f_i'^\top f_k' / \gamma)}} 
\end{align}

\begin{table}
\centering
\caption{\textbf{Model Configuration}. The `Small' model is mainly used in the analysis (\sref{sec:analysis}) while `Large-Half' model is mainly used in the results (\sref{sec:results}) for the final large-scale experiments.  `Vocab Size' is the number of token types in our model, defined by the pre-trained VQ-VAE model~\cite{ramesh2021zero}. 
}
\begin{tabular}{@{}lrrrr@{}}
\toprule
                      & Small (in \sref{sec:analysis}) & \;\;\;\;\;\;\;\;\;\;\;\; Base & Large-Half (in \sref{sec:results}) \\ \midrule
Layers                & 6                   & 12     & 24                            \\
Dimensions            & 512                 & 768    & 1024                          \\
Attention Heads       & 8                   & 12     & 16                            \\
Attention Head Dim    & 64                  & 64     & 32                            \\
MLP Intermediate Size & 2048                & 3072   & 2048                          \\
Vocab Size          & 8192                & 8192   & 8192                          \\
Params                & 29.4M               & 119.7M & 210.1M                        \\ \bottomrule
\end{tabular}
\label{tab:appen_model}
\end{table}

\begin{table}
\centering
\caption{\textbf{Training Hyperparameters}.  `Pre-Train-L' is our final Large model in \sref{sec:results} that takes a large-half model. `Pre-Train-S' is the small pre-training in analysis (\sref{sec:analysis}). `K-400' is Kinetics-400. *The batch size for pre-training is the number of samples in updating the weights. Since we use gradient accumulation, it is not correlated to the number of negative examples in \ourcl.}
\resizebox{\textwidth}{!}{%

\begin{tabular}{@{}lcc|ccccc@{}}
\toprule
                   & Pre-Train-S & Pre-Train-L & SSV2        & Diving48       & UCF101       & HMDB51       & K-400       \\ \midrule
\multicolumn{8}{l}{\footnotesize{\textit{Optimization}}} \\
Number of Epochs   & 100                 & 10                   & 22          & 50             & 50           & 50           & 30                 \\
Number of Updates  & 120K                & 12K                  & 29K         & 5.8K           & 3.7K         & 1.4K         & 48K                \\
Learning Rate      & 3e-4                & 1e-3                 & \multicolumn{5}{c}{1e-4 for small/base model, 5e-5 for large-half model} \\
Warm-Up Ratio      & 0.05                & 0.1                  & \multicolumn{5}{c}{0.1}                                                         \\
LR Decay           & \multicolumn{2}{c|}{Linear}                 & \multicolumn{5}{c}{Linear}                                                      \\
Backbone Dropout   & \multicolumn{2}{c|}{0.1}                    & \multicolumn{5}{c}{0.1}                                                         \\
Last FC Dropout    & \multicolumn{2}{c|}{-}                      & \multicolumn{5}{c}{0.0}                                                         \\
Optimizer          & \multicolumn{2}{c|}{AdamW}                  & \multicolumn{5}{c}{AdamW}                                                         \\
Batch Size         & \multicolumn{2}{c|}{1024*}                  & \multicolumn{5}{c}{128}                                                         \\
Weight-Decay       & \multicolumn{2}{c|}{0.05}                   & \multicolumn{5}{c}{0.01}                                                        \\
Adam Beta1         & \multicolumn{2}{c|}{0.9}                    & \multicolumn{5}{c}{0.9}                                                         \\
Adam Beta2         & \multicolumn{2}{c|}{0.98}                   & \multicolumn{5}{c}{0.999}                                                       \\
Adam Epsilon       & \multicolumn{2}{c|}{1e-8}                   & \multicolumn{5}{c}{1e-8}                                                        \\
Grad-Clipping Norm & \multicolumn{2}{c|}{1.0}                    & \multicolumn{5}{c}{1.0}                                                         \\ \midrule
\multicolumn{8}{l}{\footnotesize{\textit{Data Augmentation}}} \\
Color Distortion/Gray-Scale
                   & \multicolumn{2}{c|}{No}                     & \multicolumn{5}{c}{No}                                                      \\
Training Spatial Resize      
                   & \multicolumn{2}{c|}{1 (Frame Size)}                     & \multicolumn{5}{c}{2 (Frame Size, Frame Size * 1.25)}                                                      \\
Training Spatial Crops      
                   & \multicolumn{2}{c|}{1 (Center)}                & \multicolumn{5}{c}{3 (Top-Left, Center, Bottom-Right)}                                                      \\
Training Temporal Crops      
                   & \multicolumn{2}{c|}{2 (Random Uniform)}      & \multicolumn{5}{c}{1 (Random Uniform)}                                                      \\
Inference Spatial Resize      
                   & \multicolumn{2}{c|}{1 (Frame Size)}                     & \multicolumn{5}{c}{1 (Frame Size)}                                                      \\
Inference Temporal Crops     
                   & \multicolumn{2}{c|}{1 (Random Uniform)}      & \multicolumn{5}{c}{10 (Uniform)}                                                      \\ 
Training Spatial Flip      
                   & \multicolumn{2}{c|}{No}                    & \multicolumn{1}{c|}{No} & \multicolumn{4}{c}{Yes}                                                      \\
Inference Spatial Crops     
                   & \multicolumn{2}{c|}{1 (Center)}                & \multicolumn{1}{c|}{1 (Center)} & \multicolumn{4}{c}{3 (Top-Left, Center, Bottom-Right)}                                                      \\
\bottomrule
\end{tabular}
}
\label{tab:appen_hyperparam}
\end{table}

\section{Experiment Details}
\label{sec:appen_experiment}
In this section, we show our model configuration and training hyperparameters in details to support the reproducibility of our experiments.

\paragraph{Model Configuration.} Our model configuration details is shown in \tref{tab:appen_model}. Most analysis results (\sref{sec:analysis}) take `Small'  models and our final results (\sref{sec:results}) take `Large-Half' model. Other models are used in \sref{sec:model_scale_analysis}. The final `Large-Half' model halves the attention head dimension and MLP intermediate size as in \cite{child2019generating}. 
For the pre-training heads, we follow BERT to take the intermediate dimension of the token-head to be the same as the backbone's hidden dimension.
For the CLS head, we take 3 layers in MLP and 4096 intermediate dimensions. 
The output dimension is 256.
We test with different number of layers and hidden dimensions of CLS head and generally find that larger head gives better results (as in \cite{qian2020spatiotemporal, chen2021mocov3}).
This CLS head contributes to about $1\%$ pre-training computational cost overhead.

\paragraph{Training Hyperparameters.} 
We list the training hyperparameters in \tref{tab:appen_hyperparam}. 
Most of the hyperparameters are inherited from previous works to allow fair comparison and reduce tuning effort. 
For optimizer hyperparameters, we mostly follow the implementation of DALL-E~\cite{ramesh2021zero} and BERT~\cite{devlin2019bert}. 
\ssv follows the epoch number in \cite{feichtenhofer2019slowfast} and \cite{feichtenhofer2021large}.
To reduce the computational cost, we pre-extract the VQ-VAE tokens thus we employ a fixed set of spatial data augmentations.
As listed in the bottom of \tref{tab:appen_hyperparam}, we exclude any color distortion and gray scale augmentation.
We resize the video clip to the desired frame size and center-crop it during pre-training.
For downstream tasks, we resize the video clip to frame size or 1.5 times of the frame size, then crop the clip (with frame-size by frame-size spatial size) from the top-left, center, and bottom-right. 
We apply (horizontal) flip to the raw frames, thus a total of 12 spatial augmentations are extracted (12 = 2 resize $\times$ 3 crops $\times$ 2 flip/no-flip).
The only exception is \ssv.
This dataset needs to distinguish left/right motions thus we exclude the flip augmentation and only use the center crop during inference following previous works~\cite{feichtenhofer2019slowfast, feichtenhofer2021large}.
Following previous works~\cite{feichtenhofer2019slowfast}, we increase the training epochs for the non-pre-training models (in \sref{sec:pretrain_analysis}) by 4\x~for small datasets (\diving, \ucf, \hmdb) and 1.5\x~for larger datasets (\ssv, \ksmall).

\begin{table}
\centering
\caption{\textbf{Key statistics of video datasets} used in this paper. \htm is used for pre-training while others are downstream datasets. The number of training/validation examples in \hmdb and \ucf are reported for the train-val split 1.}
\resizebox{\textwidth}{!}{%
\begin{tabular}{@{}lcccccc@{}}
\toprule
                       & HowTo100M & SSV2   & Diving48 & UCF101 & HMDB51 & Kinetics-400 \\ \midrule
Training               & 1238791   & 168913 & 15027    & 9537   & 3570   & 205418       \\
Validation             & -         & 24777  & 1970     & 3783   & 1530   & 17686        \\
Number of Classes      & -         & 174    & 48       & 101    & 51     & 400          \\
Average Video Duration & 6.5min    & 4s     & 6s       & 7s     & 4s     & 10s          \\ \bottomrule
\end{tabular}%
}
\label{tab:appen_dataset}
\end{table}

\section{Dataset Details}
\label{sec:appen_dataset}
In \tref{tab:appen_dataset}, we list the key statistics of the datasets used in our paper.
\htm is our pre-training datasets that has long-duration uncurated videos.
The videos are collected from YouTube by searching key phrases thus the scale could be easily increased. 
\ssv and \ksmall are two large downstream datasets, where \ssv focuses more on the actions and \ksmall focuses more on the scenes.
\diving, \ucf, \hmdb are three small datasets.
Different from previous datasets on classifying different action types (thus might be potentially inferred from single frames), \diving studies the three stages (takeoff, flight, and entry) of competitive diving.
Thus achieving good results on \diving requires an understanding of the whole video clip.

\section{Computational Cost}
The pre-training takes around 8.9K V100 GPU hours. 
This computational cost is at the same level as ViT~\cite{dosovitskiy2020image} supervised training (5.5K hours on ImageNet-21K~\cite{russakovsky2015imagenet} and 12.8K on JFT\cite{sun2017revisiting}).
It is also at the same level of supervised training a model on Kinetics-400 dataset (6.4K for SlowFast~\cite{feichtenhofer2019slowfast}, about 5.6K for TimeSformer-L \cite{bertasius2021space}).
For fine-tuning, SSV2, Diving48, UCF101, HMDB51, and Kinetics-400 take 1K, 200, 150, 40, 2K GPU hours, respectively.
For analysis, the pre-training takes about 160 GPU hours.
Besides the final model training, energy is also spent on tuning the model and finding the best configuration.
As shown in \sref{sec:cl_analysis}, our method is more robust to the hyperparameters.

\section{Additional Analysis Results}
\label{sec:appen_analysis}

\begin{table*}
\centering
\caption{Results of \textbf{different attention-module layouts and layer-normalization positions}.  `Speed' is the normalized pre-training speed (i.e., number of samples / GPU / second). Models are pre-trained on \htm for 10 epochs. The result numbers represent \ucf accuracy.  }
\begin{tabular}{@{}ccc|cc@{}}
\toprule
        & Params & Speed & Pre-LayerNorm & Post-LayerNorm \\ \midrule
TxHxW                             & 23.1M       &  12.6         &   -            & 65.9           \\
T,HxW \cite{bertasius2021space}   & 27.1M       &  20.0         &   -            & 69.0          \\
T,H,W                             & 35.8M       &  26.4         & 69.0           & 69.6           \\
T,H|W (ours)                      & 29.4M       &  32.0         & 67.6           & 69.4           \\\bottomrule
\end{tabular}%
\label{tab:layout}
\end{table*}

\subsection{Model Architecture Comparisons}
\label{sec:appen_arch_analysis}
\paragraph{Attention Layouts.}
We here compare different alternative model architectures in Table~\ref{tab:layout}.
We first experiment with different attention layouts discussed in Sec.~\ref{sec:backbone}.
We consider the sparse attention as proposed in \cite{child2019generating} and the sequential attention blocks as in \cite{bertasius2021space}. 
The `TxHxW' model is the basic attention module that takes the flattened tokens as input (of shape T \x H \x W). 
At each layer, each token attends to all other tokens.
The `T,HxW' model separates the temporal attention and spatial attention (`Divided Space-Time' in \cite{bertasius2021space}). 
The `T,H,W' model processes three attention sequentially (`Axial Attention' in \cite{bertasius2021space}).
The `T, H|W' model is our default model that sequentially conduct temporal attention and spatial attention, where the spatial attention are parallel into the height attention and width attention.
As shown in \tref{tab:layout}, `T,H|W' reaches a balance between speed and accuracy.

\paragraph{Pre-Layer-Normalization vs. Post-Layer-Normalization.}
Besides the architectures listed above, we also consider the pre-layer-norm (used in GPT and ViT) and post-layer-norm (used in BERT) variation. 
We empirically find that post-layer-norm architecture is better for our pre-training tasks as shown in \tref{tab:layout} (comparing the last 2 columns).

\subsection{Noisy Masking for Mask-then-Predict}

Our default masking strategy replaces all masked tokens with a special \texttt{[MASK]} symbol.
We also experiment with BERT's masking strategy that only replaces $80$\% of masked tokens to the MASK symbol. 
For other tokens, $10$\% are randomly sampled from the `vocabulary' and $10$\% are kept the same.
For smaller experiments, the two masking strategies show similar results. 
However, this BERT's noisy masking strategy has lower convergence stability on the larger model pre-training.
The pre-training diverges after about $10$ epochs (out of the $100$ epochs).

\begin{table}[t]
\centering
\caption{
Impact of \textbf{masking ratio}. All models are pre-trained with only \ourmask task. 
\#Blocks is the number of masking blocks.
Default setup is underlined. 
}
\begin{tabular}{@{}c|cc|cc@{}}
\toprule
Strategy & \#Blocks & Ratio & Mask-Accu.$\uparrow$                  & UCF101  $\uparrow$        \\ \midrule
Block            & 4              & 11.9\%        &  17.9                       & 66.8            \\
\underline{Block}            & \underline{5}              & \underline{14.5\%}        &  17.6                       & 68.3           \\
Block            & 6              & 17.0\%        &  17.3                       & 67.3            \\ 
i.i.d.           & -              & 11.9\%        &  25.6                       & 64.5                \\
i.i.d.           & -              & 14.5\%        &  24.3                       & 63.5           \\
i.i.d.           & -              & 17.0\%        &  24.0                       & 64.0           \\ \bottomrule
\end{tabular}%
\label{tab:appen_masking_ratio}
\end{table}

\subsection{Masking Ratio for Block-Masking and I.I.D. Masking}
We test the effect of different masking ratios.
In the main text, we control the number of blocks for block masking.
In \tref{tab:appen_masking_ratio}, we here also show the results of matched masking ratio for i.i.d. masking for completeness.
Empirically, the result differences among various masking ratios are marginal and the original BERT's $15$\% masking ratio (with roughly $5$ masking blocks) works slightly better.
Thus we always select the number of mask blocks whose induced masking ratio is closest to $15$\%.
For all masking ratios, block masking shows significantly better results than the i.i.d. masking.

\begin{table}[t]
\centering
\caption{ Impact of \textbf{\ourcl loss weight $\alpha$}. Default setup is underlined. }
\begin{tabular}{l|ccc}
\toprule
$\alpha$           & Mask-Accu.$\uparrow$ & CL-Loss$\downarrow$ & UCF101$\uparrow$ \\ \midrule
0.0                & 17.6      & -       & 68.3  \\
0.5                & 17.5      & 1.07    & 70.2  \\
\underline{1.0}                & 17.2      & 1.06    & 69.4  \\
2.0                & 16.9      & 1.05    & 68.0  \\
$\infty^*$         & -         & 1.07    & 57.1  \\
\bottomrule
\end{tabular}%
\label{tab:loss_ratio}
\end{table}

\subsection{Impact of Contrastive Learning Loss Weight}
In \tref{tab:loss_ratio}, we show the impact of loss weight $\alpha$
(see Sec.~\ref{sec:objective}).
Since the loss have been calibrated by multiplying the temperature, $\alpha\mbox{=}1$ shows stable results and $\alpha\mbox{=}0.5$ is slightly better. 
Setting $\alpha\mbox{=}2.0$ will let the model to focus mostly on \ourcl task and its result is worse than pure \ourmask pre-training (i.e., $\alpha\mbox{=}0.0$).
We also list the pure \ourcl pre-training results here (denoted as $\alpha\mbox{=}\infty^*$ but it excludes the \ourmask loss and set $\alpha\mbox{=}1.0$) for reference.

\section{Visualizations}
\label{sec:appen_visualization}

\begin{figure}[t]
\vskip 0.1in
\begin{center}
\includegraphics[
                width=0.8\columnwidth,
                ]{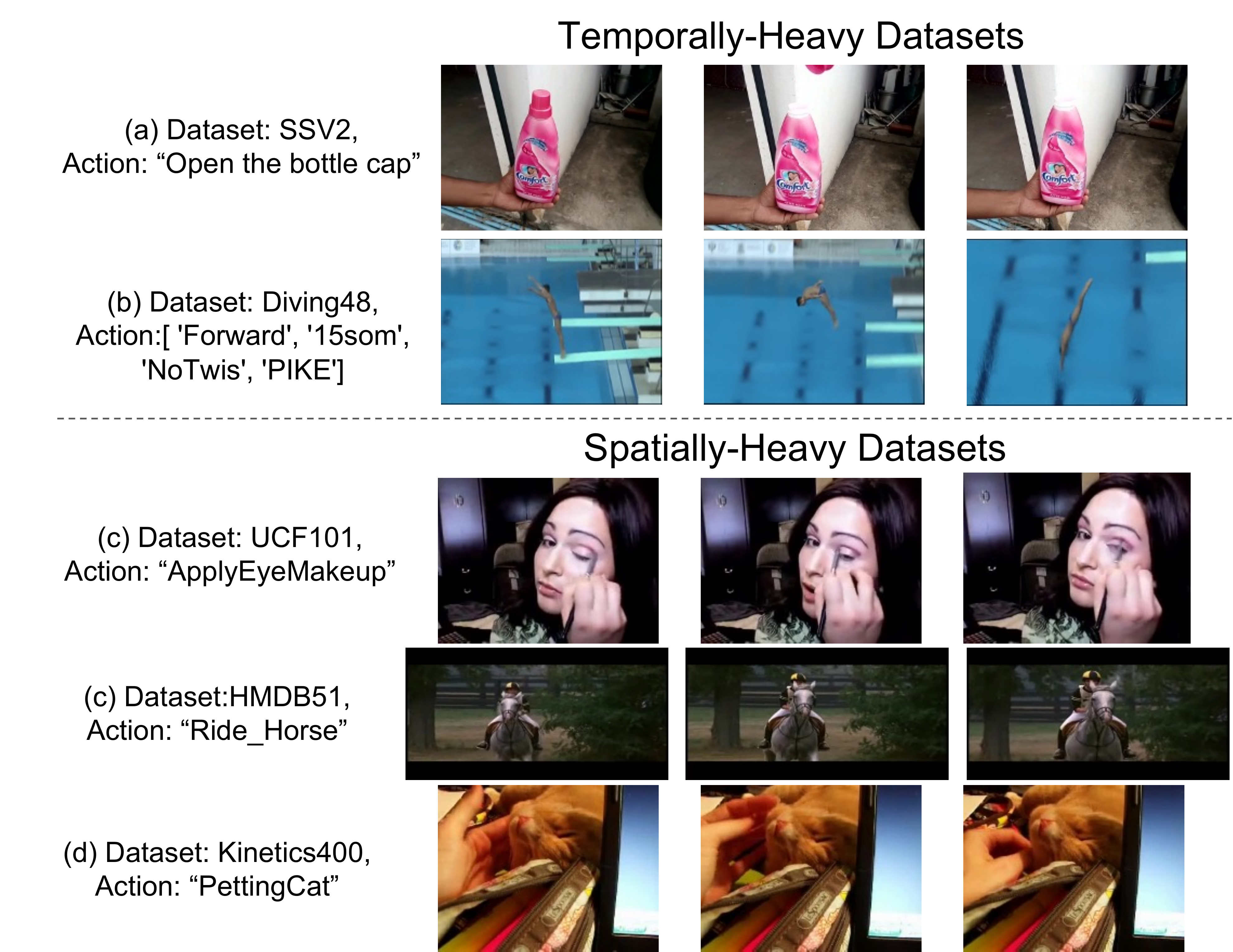}
                \vspace{-6pt}
\caption{
Data samples from temporally-heavy and spatially-heavy datasets.
While temporally-heavy datasets need the temporal information to make decisions, most actions in spatially-heavy datasets could be inferred from single frames.
}
\label{fig:dataset_vis}
\end{center}
\vspace{-10pt}
\end{figure}

\subsection{Temporally-Heavy vs. Spatially-Heavy Datasets}
\label{sec:append_dataset_visualization}
We illustrate the differences between temporally-heavy and spatially-heavy datasets in \fref{fig:dataset_vis}.
We here show equally-distributed frames from the video and the label of the video clip.
Note that we do not cherry-pick the data but aim for showing the nature of each dataset.
Overall, understanding in temporally-heavy datasets needs temporal modeling, whereas the action labels of spatially-heavy datasets could be inferred from a single frame.
To understand the SSV2\cite{goyal2017something} example in \fref{fig:dataset_vis}.(a), the model needs to understand the causality, i.e., the order of the frames decides the action label.
In \fref{fig:dataset_vis}.(b), the competitive diving dataset Diving48~\cite{li2018resound} also requires considering nearly all frames to make the decision.
However, for the spatially-heavy datasets (UCF101~\cite{soomro2012ucf101}, HMDB51~\cite{kuehne2011hmdb}, Kinetics-400~\cite{carreira2017quo}), the action label could be inferred from any single sampled frame. 
These observations result in the pretty high frame-level accuracy (i.e., not modeling temporal interactions) in \sref{sec:results}.

\begin{figure}[t]
\vskip 0.1in
\begin{center}
\includegraphics[
                width=0.95\columnwidth,
                ]{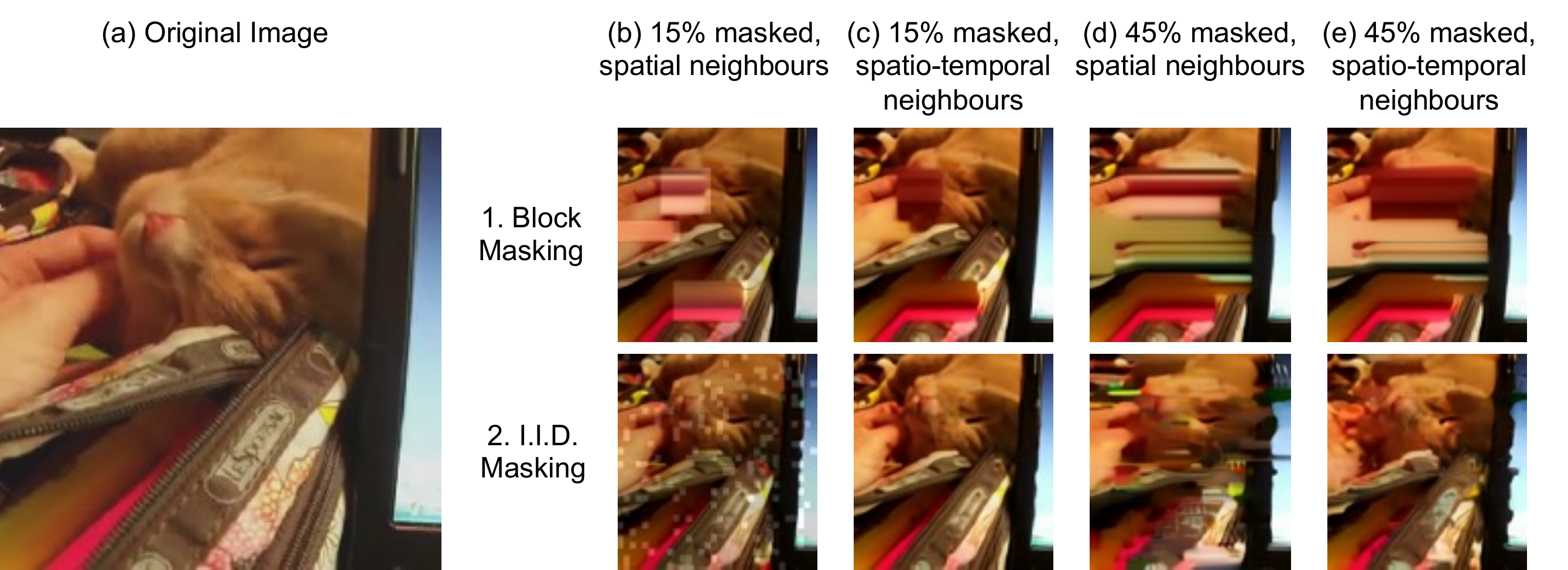}
                \vspace{-6pt}
\caption{
Nearest-neighbour reconstruction of  \textit{block masking} and \textit{i.i.d masking}.
We mask tokens at different ratios and reconstruct them by simply copying their spatial or spatio-temporal neighbours.
Even under heavy masking (e.g., 45\% masked), this simple reconstruction strategy still yields a reasonable results for \textit{i.i.d masking}, e.g., we can easily recognize the action `petting cat' from the reconstructed images, especially the one reconstructed from spatio-temporal neighbours. However, this becomes significantly more difficult when using \textit{block masking}.
}
\label{fig:block_vs_iid}
\end{center}
\vspace{-10pt}
\end{figure}

\subsection{Masking Strategy Comparisons} 
\label{sec:append_mask_visualization}
We propose to use block masking (in \sref{sec:mask_task}) since i.i.d. masking  may lead to trivial solutions for the \ourmask task given the strong localities in videos.
We illustrate this point in \fref{fig:block_vs_iid} with a simple copy-paste reconstruction method. 
Specifically, after masking, we first replace the masked tokens with their \textit{nearest visible neighbours} (i.e., the unmasked token that has the shortest distance in spatial or spatio-temporal domain), and then forward the reconstructed tokens to the VQ-VAE decoder to generate the RGB images.
For the default 15\% masking ratio, i.i.d. masking is recoverable while block masking causes striped noisy patterns.\footnote{We highlight the masking region in \fref{fig:block_vs_iid}(b) and show the raw RGB images in other cases.}
We also test with the extreme case of masking 45\% tokens (in \fref{fig:block_vs_iid} (d), (e)).
The block-masked images are hard to reconstruct, however, some objects in reconstructed images from i.i.d. masking are still recognizable.
When comparing images under the same masking strategy, recovered images using spatio-temporal neighbours is better than using only spatial neighbours, especially when comparing the images under 45\% i.i.d. masking (i.e., (d).2 and (e).2 in \fref{fig:block_vs_iid}).
Overall, these results indicate that using i.i.d. masking in \ourmask task has a potential trivial solution by copying the neighbourhood, while block-masking resolves this issue.

\begin{figure}[t]
\vskip 0.1in
\begin{center}
\includegraphics[
                width=0.95\columnwidth,
                ]{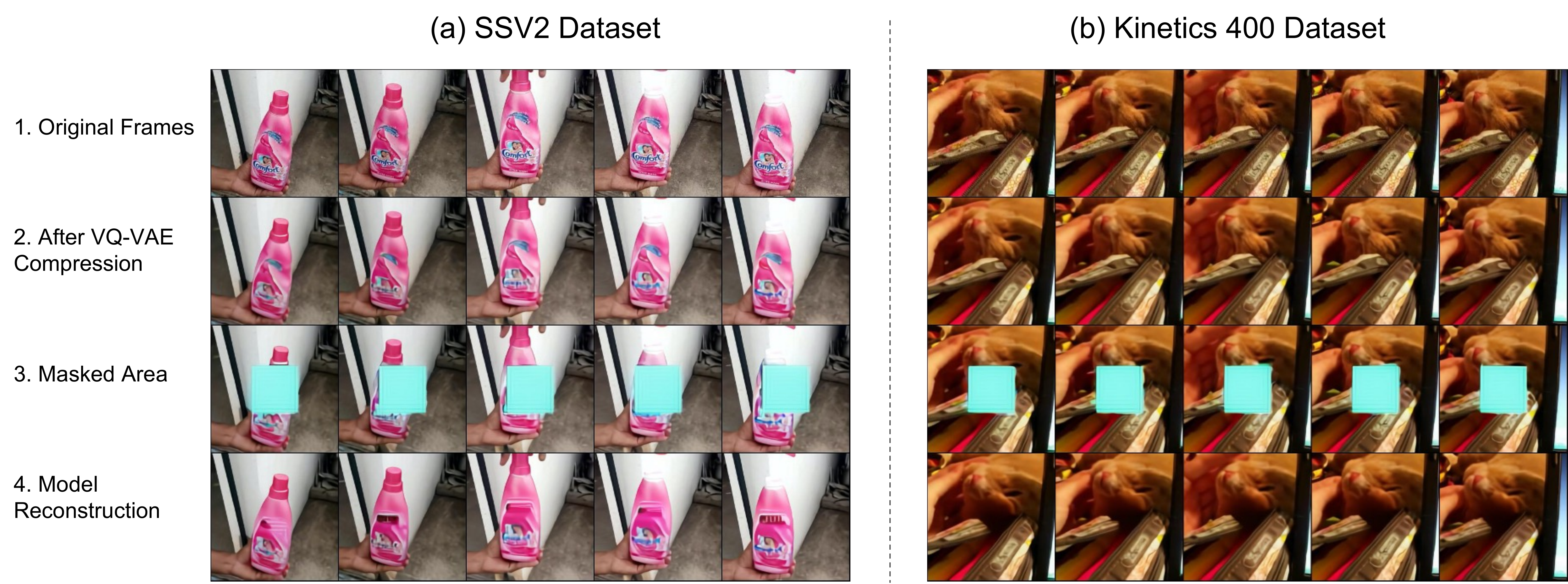}
\caption{
Masked-token model reconstruction for SSV2 and Kinetics-400 datasets. Comparing 1. and 4., our model could redraw temporally-consistent and spatially-plausible patches for the masked regions.
}
\label{fig:reconstruction}
\end{center}
\end{figure}

\subsection{Model Reconstruction}
\label{sec:append_recon_visualization}
Since our model is trained with \ourmask task, it is able to reconstruct masked tokens.
In this section, we showcase the reconstructed video frames by our final model (i.e., 24 layers, 1024 dimensions, 5 clip length, and 256 frame size).
As shown in \fref{fig:reconstruction}, we provide two examples from the SSV2 and Kinetics-400 dataset.
We uniformly sample 5 consecutive frames from the video at 1 frame per second.
We show the original frames in the first rows (\fref{fig:reconstruction}.(a).1, (b).1).
As illustrated in \sref{sec:append_dataset_visualization}, the temporally-heavy SSV2 dataset has object motions between frames while the spatially-heavy Kinetics-400 dataset has almost static frames.
In the second rows, we show the images after VQ-VAE compression.
To do this, we first use VQ-VAE encoder to encode the images, and then use VQ-VAE decoder to reconstruct the images, without any corruptions in between.
We see that there is some information loss caused by the VQ-VAE compression (e.g., the text `comfort' in \fref{fig:reconstruction}.(a).1).  
It potentially contributes to relative lower results on spatially-heavy datasets.
In the third and fourth rows, we illustrate the masked area and the prediction from our model.
As shown in \fref{fig:reconstruction}.(a).4, our model could faithfully redraw the shape and texture of the object.
As shown in \fref{fig:reconstruction}.(b).4, the shape of the cat's head is pretty similar to the original frames while the shading is different (but still consistent in different frames).